\PassOptionsToPackage{unicode}{hyperref}
\PassOptionsToPackage{hyphens}{url}
\PassOptionsToPackage{dvipsnames,svgnames,x11names}{xcolor}
\documentclass[
  10pt,
  letterpaper,
  twocolumn]{article}
\usepackage{xcolor}
\usepackage[letterpaper,width=6.75in,height=9.25in,centering]{geometry}
\usepackage{amsmath,amssymb}
\usepackage{iftex}
\ifPDFTeX
  \usepackage[T1]{fontenc}
  \usepackage[utf8]{inputenc}
  \usepackage{textcomp} %
\else %
  \usepackage{unicode-math} %
  \defaultfontfeatures{Scale=MatchLowercase}
  \defaultfontfeatures[\rmfamily]{Ligatures=TeX,Scale=1}
\fi
\usepackage{lmodern}
\ifPDFTeX\else
\fi
\IfFileExists{upquote.sty}{\usepackage{upquote}}{}
\IfFileExists{microtype.sty}{%
  \usepackage[]{microtype}
  \UseMicrotypeSet[protrusion]{basicmath} %
}{}
\usepackage{setspace}
\makeatletter
\@ifundefined{KOMAClassName}{%
  \IfFileExists{parskip.sty}{%
    \usepackage{parskip}
  }{%
    \setlength{\parindent}{0pt}
    \setlength{\parskip}{6pt plus 2pt minus 1pt}}
}{%
  \KOMAoptions{parskip=half}}
\makeatother
\usepackage{graphicx}
\makeatletter
\newsavebox\pandoc@box
\newcommand*\pandocbounded[1]{%
  \sbox\pandoc@box{#1}%
  \Gscale@div\@tempa{\textheight}{\dimexpr\ht\pandoc@box+\dp\pandoc@box\relax}%
  \Gscale@div\@tempb{\linewidth}{\wd\pandoc@box}%
  \ifdim\@tempb\p@<\@tempa\p@\let\@tempa\@tempb\fi%
  \ifdim\@tempa\p@<\p@\scalebox{\@tempa}{\usebox\pandoc@box}%
  \else\usebox{\pandoc@box}%
  \fi%
}
\def\fps@figure{htbp}
\makeatother
\providecommand{\tightlist}{%
  \setlength{\itemsep}{0pt}\setlength{\parskip}{0pt}}
\usepackage[style=apa,backend=biber,natbib=true]{biblatex}
\usepackage{amsmath}
\usepackage{amssymb}
\usepackage{amsthm}
\usepackage{mathtools}
\usepackage{bm}

\usepackage{graphicx}
\usepackage{caption}
\usepackage{subcaption}
\usepackage{float}
\usepackage{wrapfig}

\usepackage{setspace}
\usepackage{indentfirst}

\usepackage{microtype}

\usepackage{booktabs}
\usepackage{longtable}
\usepackage{array}
\usepackage{multirow}
\usepackage{threeparttable}

\usepackage{hyperref}
\hypersetup{
  colorlinks=true,
  linkcolor=black,
  citecolor=blue,
  urlcolor=blue,
  pdfborder={0 0 0}
}

\newtheorem{theorem}{Theorem}[section]

\newtheorem{definition}[theorem]{Definition}

\usepackage{booktabs}
\usepackage{longtable}
\usepackage{array}
\usepackage{multirow}
\usepackage{wrapfig}
\usepackage{float}
\usepackage{colortbl}
\usepackage{pdflscape}
\usepackage{tabu}
\usepackage{threeparttable}
\usepackage{threeparttablex}
\usepackage[normalem]{ulem}
\usepackage{makecell}
\usepackage{xcolor}
\usepackage{bookmark}
\IfFileExists{xurl.sty}{\usepackage{xurl}}{} %
\hypersetup{
  pdftitle={Persistent Convolution: A Topological Framework for AI Alignment Testing and Semantic Space Characterization},
  pdfauthor={Tyler Ashoff; Jordan Rodu},
  colorlinks=true,
  linkcolor={blue},
  filecolor={Maroon},
  citecolor={blue},
  urlcolor={blue},
  pdfcreator={LaTeX via pandoc}}

\title{Persistent Convolution: A Topological Framework for AI Alignment
Testing and Semantic Space Characterization}
\author{Tyler
Ashoff\thanks{Department of Statistics, University of Virginia. Correspondence: tlashoff@gmail.com. This work is derived in part from the first author's doctoral dissertation \parencite{ashoff2026dissertation}.} \and Jordan
Rodu\thanks{Department of Statistics, University of Virginia}}
\date{July 2026}

\begin{document}
\maketitle

\setstretch{1}
Modern opaque AI models prize performance over interpretability, which
makes testing difficult. However, formal statistical tests conducted on
a model's embedding space can provide robust characterizations of
semantic structure, concept separation, and knowledge graph alignment.
Model developers would benefit from a model comparison technique that
leverages human-curated knowledge structures to test alignment. The
scale of the input space for even relatively simple tasks motivates the
need for alignment checks that augment standard outcome reasoning. This
work develops and demonstrates a topology-based multi-modal alignment
test to make deployment, selection, and comparison of opaque models more
interpretable. These methods also offer an intuitive connection to
possibility theory and a unified decision theoretic framework from data
to
deployment.\footnote{Implementation available at \href{https://github.com/tylerashoff/persiscope}{github.com/tylerashoff/persiscope} and on PyPI as \texttt{persiscope}.}

\begin{center}\rule{0.5\linewidth}{0.5pt}\end{center}

\section{Introduction}\label{introduction}

\subsection{Motivation}\label{motivation}

Machine learning and artificial intelligence (AI) models are notoriously
opaque. They often provide useful predictions, but their complexity
restricts their interpretability. Users of these models may not be
interested in how predictions are generated, but policy-makers and model
developers need more insight. Model evaluations should allow
decision-makers to grade a model's robustness to domain shift. Current
evaluation methods try to address these goals, but fall short.
Traditional statistical modeling relies on interpretable coefficients
and distribution parameters to justify the structure of a model and
ensure that test results are both meaningful and generalizable. AI model
structures are clever, but complex, which makes them difficult to
interpret. So, AI evaluations grade robustness with test-set
performance, substituting explanation for extrapolation and sense for
scale. The breadth of modern AI applications is enormous. This reliance
on multiple-choice and crowdsourced testing leaves developers without a
rigorous tool to assess a model's reliability.

``Double descent'' \autocite{reconciling-double-descent-19} challenges
the classic bias-variance trade-off of modeling and suggests modern AI
models have moved beyond an under-fit or over-fit regime into an
interpolation regime. Classically, model developers balance the number
of parameters in their models so the model captures information from the
training set but retains the ability to generalize to unseen data.
Double descent suggests that if enough data is gathered to effectively
cover the input and output space of the models, any task requires only
slight variations from the training data or ``interpolation'' between
densely packed data points. A second drop in test error appears once a
model enters the interpolation regime, which gives the work its name.
This is a fundamental shift in the purpose of model building. Classical
model development cares about model performance, but also values the
meaning derived from model parameters and the insight this meaning gives
to the underlying data generation process. Modern opaque AI models prize
performance over interpretability, which makes testing these models
difficult.

As use cases become more general, generating exhaustive inputs becomes
infeasible. The embedding space of these models offers richer
information about how inputs and outputs are represented internally.
Model developers often inspect this embedding space visually but formal
tests can be conducted on a model's embedding space, to pull model
evaluation away from ad hoc testing and towards statistical rigor.

The scale, complexity, and popularity of modern deep learning models
makes them a useful example for this work. However, these issues also
plague AI models in other domains. In many ways, it is the smaller
models, operating in data-scarce domains that would gain the most from a
more rigorous evaluation framework. Large models benefit from abundant
data to ingest and many eyes watching them. They rely on this attention
to cover the input space and catch errors that internal evaluators miss.
Bespoke models and niche or data-limited domains may not be so lucky. In
domains where data is difficult or expensive to collect, test-set scores
may become saturated, making model selection based on performance alone
difficult. For these situations, model developers would benefit from a
model comparison technique that leverages human-curated knowledge
structures to test alignment.

This work focuses on vision models since they are well studied and the
data are intuitive. However, since these methods focus on the semantic
structure, they are domain agnostic and multi-modal.

\subsection{Existing Work}\label{existing-work}

Work on AI alignment, mechanistic interpretability, and semantic
structure share common goals: characterize how models ``think'', learn
to effectively manipulate model behavior, and govern the output of
models robustly and at scale.

Existing work has used embedding spaces to evaluate models. The
``Embedding Comparator'' \autocite{boggust2022embedding} tool uses local
similarity metrics to interactively surface inputs whose embedded
representation is dissimilar between models. By comparing the embeddings
generated by a model's training data to the embedding of new data, ALICE
\autocite{rajendran2019accurate} calculates a ``competency'' score to
signal the model's familiarity with incoming data. The Conceptualizing
Embedding Spaces (CES) \autocite{simhi2023interpreting} algorithm maps
embedded representations to a new space grouping concepts based on a
given hierarchical ontology. The research most similar to this work is
Data Kernel Perspective Spaces (DKPS) \autocite{dkps25} which re-embeds
model output, creates a reduced dimensional representation, and builds a
discriminator model to separate data points in the reduced space.
However, these methods do not formally compare semantic spaces using a
consistent characterization of the model's own latent space nor do they
show how to test alignment with human knowledge structures.

Research into the shape of these semantic spaces and how to align these
structures with human preferences is the main motivation for this work.
Rather than finding individual concept spaces, this work tries to
characterize the underlying topology of the entire space and create
rigorously comparable representations. To lean into the
anthropomorphization of these models, this work seeks to capture and
compare model personalities by aggregating the models' value and
behavior vectors into a robust representation.

\subsection{Task-based Testing in the Real
World}\label{task-based-testing-in-the-real-world}

Model characterizations that do not depend on testing every possible
input are needed. While synthetic data may play a role in testing these
models, it is not a replacement for real-world testing. Unfortunately,
real-world testing can be expensive and time consuming. When researching
how long it would take to exhaustively test autonomous cars, where fatal
or injurious failures occur with low frequency, researchers determined:

\begin{singlespace}
\begin{quote}
\textit{"Under even aggressive testing assumptions, existing fleets would take tens and sometimes hundreds of years to drive these miles — an impossible proposition if the aim is to demonstrate their performance prior to releasing them on the roads for consumer use."}

\hfill -- RAND Corporation, \textcite{drivingtosaftey}
\end{quote}
\end{singlespace}

This work does not propose a complete solution to this problem, as it is
still fundamentally limited by the number and diversity of examples
provided. However, the scale of the input space for even relatively
simple tasks motivates the need for alignment checks that augment
task-based testing. Formal methods for testing concept separation and
knowledge graph alignment can help towards this goal.

This work uses a set of tools from topology and statistics used to
conduct these formal tests. First, topological representations are
defined which capture information about the embeddings' structure in
clean functional forms: persistence landscapes and persistence
silhouettes. Then, standard permutation methods are used to formally
test the similarity of these topological representations between
different embedding techniques. These two steps enable a
statistically-grounded alignment test between embeddings of any
dimensionality.

The main goal of this work is to develop and demonstrate a
topology-based multi-modal alignment test to make deployment, selection,
and comparison of opaque models more interpretable. The main
contributions of this work are:

\begin{itemize}
\tightlist
\item
  Adjusted topological representations to capture changes in semantic
  connectivity
\item
  Formal alignment test between semantic space and behavior preferences
\item
  Multi-modal and dimensionally agnostic semantic comparisons
\item
  Metric-guided model training and selection over varying preferences
\end{itemize}

\begin{center}\rule{0.5\linewidth}{0.5pt}\end{center}

\section{The Persistent Convolution
Framework}\label{the-persistent-convolution-framework}

This work leverages formal hypothesis tests on topological features to
characterize embedding spaces, which enables comparisons at various
scales, from local behavior to overall global structure. These
topological tools for characterizing structure is where this work starts
and alignment tests are built to check the similarity of knowledge
representations.

\begin{figure*}[!t]
\centering

\begin{center}\includegraphics[width=1\linewidth]{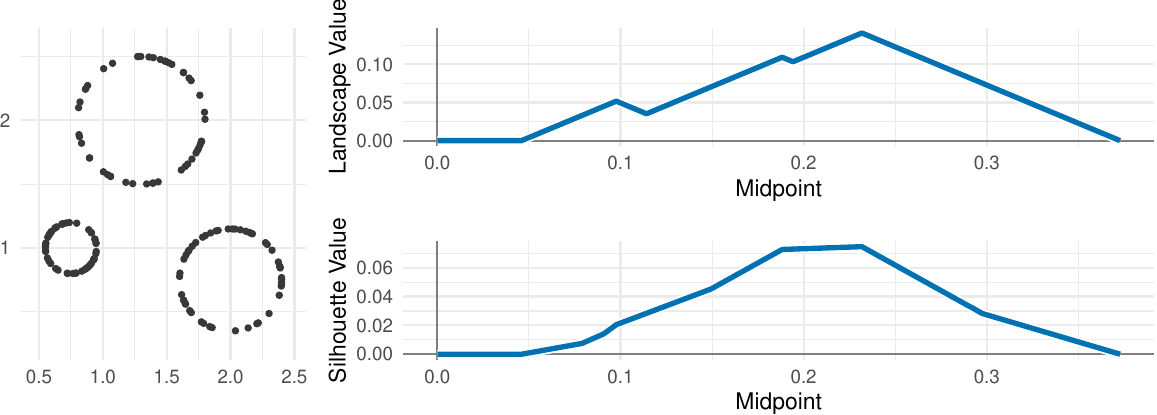} \end{center}
\caption{Two dimensional embedding and the associated persistence landscape and persistence silhouette based the $1^{\mathrm{st}}$ homology group}
\label{fig:tda-emb-to-rep}
\end{figure*}

Figure \ref{fig:tda-emb-to-rep} shows a two dimensional embedding space
with three rings and two topological representations of this embedding:
a persistence landscape and a persistence silhouette. The following
sections step through how these representations are developed and how
they are used to compare embedding spaces. These tools will result in a
rigorous two-sample homogeneity test between the embeddings.

Importantly, these tests are not restricted to embedding spaces with the
same dimension. The technique takes as input, two full connected graphs
and builds two topological representations, so it is not limited to
comparing spaces within a single model or family of models. By focusing
on connectivity with graphs, comparisons are possible between models of
arbitrary dimension, including formal knowledge structures. These
structures may be formal ontologies or intuitive two dimensional plots
where people can plot the relationship between concepts visually. In
this way, the tool becomes an alignment check against interpretable
encodings rather than only an interesting characterization of high
dimensional vector spaces. Defining these ``baselines'' is a focus of
the results section, while this section focuses on formally developing
the tools.

\subsection{Embedded Data to Persistence
Diagrams}\label{embedded-data-to-persistence-diagrams}

\begin{figure*}[!t]
\centering

\begin{center}\includegraphics[width=1.0\textwidth]{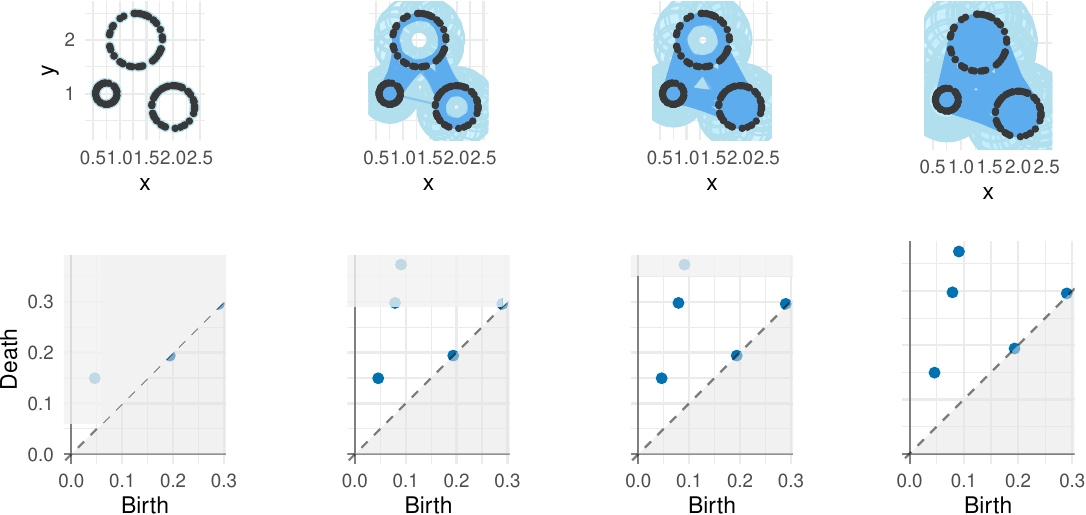} \end{center}
\caption{Demonstration of how a filtration of an embedding space generates a persistence diagram based on the $1^{\mathrm{st}}$ homology group}
\label{fig:filtration-example}
\end{figure*}

The first step in building the landscapes and silhouettes is to create a
persistence diagram from the embedding space. Figure
\ref{fig:filtration-example} shows this process. The top plots show a
filtration of the embedding space. Neighborhoods are grown around each
data point at the same rate while keeping track of the radius of the
neighborhoods. When the neighborhoods of two points intersect, an edge
is drawn between those points. This process continues until each point
is connected to every other point. The persistence diagrams in the
bottom row capture the evolution of this connection process.

Persistence diagrams are designed to identify different types of
structures called ``homology groups''. Unintuitively, the indexing
starts at 0, so the \(0^{\mathrm{th}}\) homology group tracks how
individual components merge into groups, the \(1^{\mathrm{st}}\)
homology group tracks when cycles or loops form, the \(2^{\mathrm{nd}}\)
homology group tracks voids in spaces with more than two dimensions, and
higher order groups track increasingly more abstract topological
features in high dimensional space. This work focuses on the
\(0^{\mathrm{th}}\) homology group in the results but uses the
\(1^{\mathrm{st}}\) homology group for demonstration since loops are
visually more intuitive than connected components and the overall
process is nearly identical.

Figure \ref{fig:filtration-example} tracks the evolution of the
\(1^{\mathrm{st}}\) homology group, or the loops. Loops are ``born''
when a cycle is first created and they ``die'' when the points are
connected across the feature. Those (birth, death) persistence pairs are
the coordinates on the persistence diagram plot. On the left, the
smallest ring has made a cycle but the other two rings have not been
connected. In the second panel, the smallest ring has closed out, so the
persistence pair has been plotted as the left most point. The other two
rings have also been connected and are on the verge of closing out, but
the death time is not known to plot the feature in the diagram. The
final two panes show each of the two largest rings closing and being
plotted as the two highest points.

The additional features also show up in this diagram. The two points
near the diagonal represent small features that do not last very long.
In the second pane, a small cycle is created in between the three large
loops. Since it dies quickly after it is born, the persistence
coordinate falls close to the diagonal. Features that have a larger
difference between birth and death times lift off of the diagonal and
are said to ``persist'' longer, giving the diagram its name. A
consequence of this pairing also means that no point can exist below the
diagonal, which will be important in the following transformations.

Also, an important side note is that if multiple disjoint features
happen to be born and die at the same times, the point associated with
this persistence pair has a multiplicity greater than one. This is
important in the construction of the silhouettes.

\subsubsection{Homology Groups and Betti
Numbers}\label{homology-groups-and-betti-numbers}

The \(p^{th}\) homology group captures the evolution of the topological
features (e.g.~connected components(\(p=0\)), loops(\(p=1\)),
voids(\(p=2\)))\autocite{TDA}. It is defined as the quotient vector
space \[H_p(K) = Z_p(K)/B_p(K)\]

On a simplicial complex \(K\), \(C_p\) is a group of \(p\)-chains,
\(\delta_p: C_p\to C_{p-1}\) is the boundary homomorphism,
\(Z_p(K) = ker(\delta_p)\) is the group of \(p\)-cycles, and
\(B_p(K)= im(\delta_{p+1})\) is the group of \(p\)-boundaries. The Betti
number, \(\beta_p = rank(H_p)\), tracks the count of these features.

\subsubsection{Filtrations}\label{filtrations}

\begin{singlespace}
\begin{definition}[\cite{stab-pm}]
\label{def:filtrations}
$\mathbb{X}_{sub}^f$ is the sublevel set filtration of $(X, f)$ where $f: X \to \mathbb{R}$ and $X$ is a topological space.
The sublevel sets are $$X^t = \left(X,f\right)^t = \left\{ x\in X | f(x) \leq t\right\}$$ where the inclusion maps $\left\{i_s^t: X^s \to X^t | s \leq t\right\}$ satisfy $i_s^t \circ i_r^t$ for $r\leq s \leq t$ and $i_t^t$ is the identity on $X^t$.
\end{definition}
\end{singlespace}

For \(\mathbb{X}\), a subset of a compact metric space \((M, \rho)\),
the families of Rips-Vietoris complexes
\(\left(Rips_r(\mathbb{X})\right)_{r\in \mathbb{R}}\) and \v{C}ech
complexes \(\left(Cech_r(\mathbb{X})\right)_{r\in \mathbb{R}}\) are
filtrations.

\begin{singlespace}
\begin{theorem}[Nerve Theorem, \cite{tda-intro}]
\label{nerve-theorem}
Let $\mathcal{U}=\left(u_i\right)_{i\in I}$ be a cover of the topological space $X$ by open sets s.t. the intersection of any subcollection of the $u_i$'s is either empty or countable. Then, $X$ and the nerve $C(\mathcal{U})$ are homotopy equivalent.
\end{theorem}
\end{singlespace}

If \(\mathbb{X}\) is a point cloud in \(\mathbb{R}^d\), the Nerve
Theorem shows that the \v{C}ech complex encodes the topology of the
whole family of union of balls
\(\mathbb{X}^r = \bigcup_{x\in\mathbb{X}} B(x, r)\) as \(r\to \infty\)
for \(r \geq 0\).

\subsubsection{Persistence Module}\label{persistence-module}

\begin{singlespace}
\begin{definition}[\cite{stab-pm}]
\label{def:pers-module}
A persistence module $\mathbb{V}$ over $\mathbb{R}$ is an indexed family of vector spaces $\left(V_t | t\in \mathbb{R}\right)$ and a doubly-indexed family of linear maps $\left(v_s^t: V_s \to V_t | s \leq t \right)$ which satisfy the composition law $v_s^t \circ v_r^s = v_r^t$ where $r \leq s \leq t$ and $v_t^t$ is the identity map on $V_t$.
\end{definition}
\end{singlespace}

To obtain a persistence module, apply \(p^{th}\) singular homology
group\(, H_p(\cdot)\), a functor from topological to vector space. Set
\(V_t = H_p\left(x^t\right)\) and
\(v_s^t = H_p\left(i_s^t\right): H_p\left(x^s\right) \to H_p\left(x^t\right)\)
to obtain a persistence module
\[\mathbb{V} = H_p\left(\mathbb{X}^f_{sub}\right)\]

Given any \(f: X \to \mathbb{R}\) and \(a\in\mathbb{R}\),
\[\mathbb{V}\left(f\right)(a) = H_p\left(f^{-1}\left((\infty, a]\right)\right)\]
where \(\mathbb{V}\left(f\right)(a\leq b)\) is induced by inclusion
\autocite{bubenik2015statistical}.

Betti numbers can be seen in terms of the image of a persistence module,
\[\beta^{b, d} = dim\left(img\left(\mathbb{V}\left(f\right)(b\leq d)\right)\right)\]
where \(f\) is the minimum distance to a finite set of points. Stable
persistence diagrams can be constructed from q-tame persistence modules
\autocite{stab-of-tame} \autocite{stab-pm}.

\subsubsection{Persistence Diagrams}\label{persistence-diagrams}

\begin{singlespace}
\begin{definition}[\cite{pd-stability}]
\label{def:pd-critical-values}
Let $f$ be a real function on $X$. $r\in \mathbb{R}$ is a homological critical value of $f$ if $\exists$ $p\in\mathbb{Z}$ s.t. $\forall$ $\epsilon > 0$ the map $H_p\left(f^{−1}\left((−\infty,r− \epsilon]\right)\right) \to H_p\left(f^{−1}\left((−\infty, r+ \epsilon]\right)\right)$ is not an isomorphism.
\end{definition}
\end{singlespace}

For each subcomplex \(K^r\) of the finite simplicial complex \(K\),
identify ``critical values'' of \(r\) where the complex changes. Then
for \(r_1 < r_2 < ... < r_n\), the finite diagram
\[H_p\left(K^{r_1}\right) \to H_p\left(K^{r_2}\right) \to \cdots \to H_p\left(K^{r_n}\right)\]
contains all the information in the persistence module
\autocite{stab-pm}.

\begin{singlespace}
\begin{definition}[\cite{pd-stability}]
\label{def:pers-diagram}
The persistence diagram $D\left(f\right)\subset \overline{\mathbb{R}}^2$ of $f$ is the multiset of points $\left\{(r_i, r_j)\right\}_{0\leq i < j \leq n+1}$ union all points on the diagonal counted with infinite multiplicity.
\end{definition}
\end{singlespace}

Persistence diagrams are created by tracking how the homology groups
evolve over the filtration. The off-diagonal critical values correspond
to either the emergence of new homological features or merging of
existing features. These events are referred to as ``birth'' and
``death'' of features respectively and create the multiset of real
intervals, ``persistence pairs'', \({\left(b_i, d_i\right)}_{i\in I}\)
where I is a finite set.

By convention, the merging of two features follows the ``elder rule'',
where the older feature survives and the younger feature dies.

\begin{figure*}[!t]
\centering

\begin{center}\includegraphics[width=1.0\textwidth]{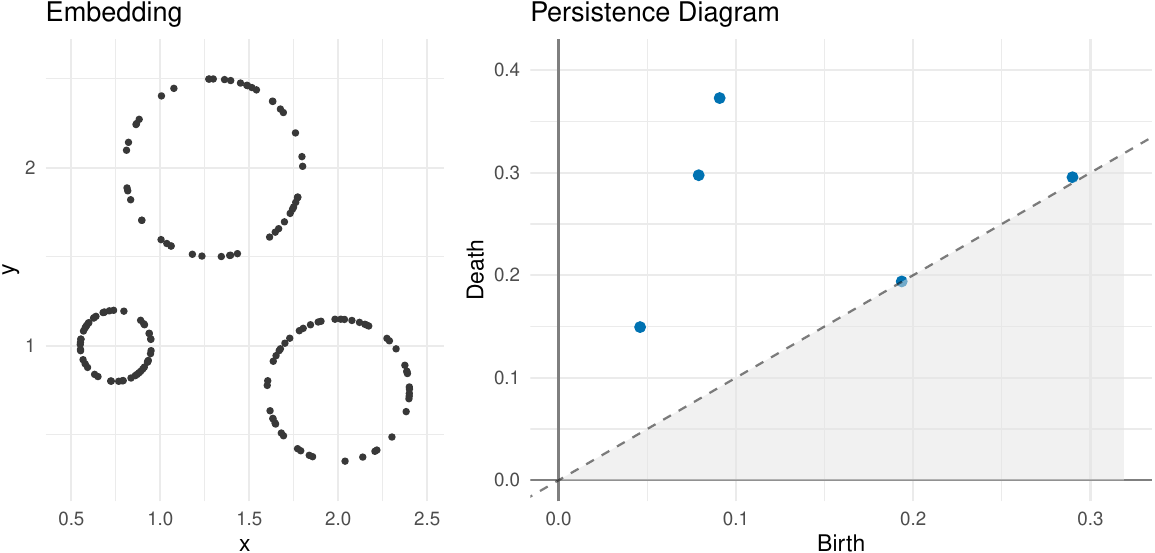} \end{center}
\caption{Persistence diagram based on an embedding space and the $1^{\mathrm{st}}$ homology group}
\label{fig:pd}
\end{figure*}

\subsubsection{Homology Dimensions}\label{homology-dimensions}

The \(1^{\mathrm{st}}\) homology dimension (\(H1\)) which tracks loops
is used while introducing the persistence diagrams above. However,
comparing how individual points cluster may be more interesting than how
many loops exist in an embedding space. The \(0^{\mathrm{th}}\) homology
dimension (\(H0\)) tracks this clustering, it is focused on tracking
connected components as the neighborhoods grow during the filtration.

\begin{figure*}[!t]
\centering

\begin{center}\includegraphics[width=1.0\textwidth]{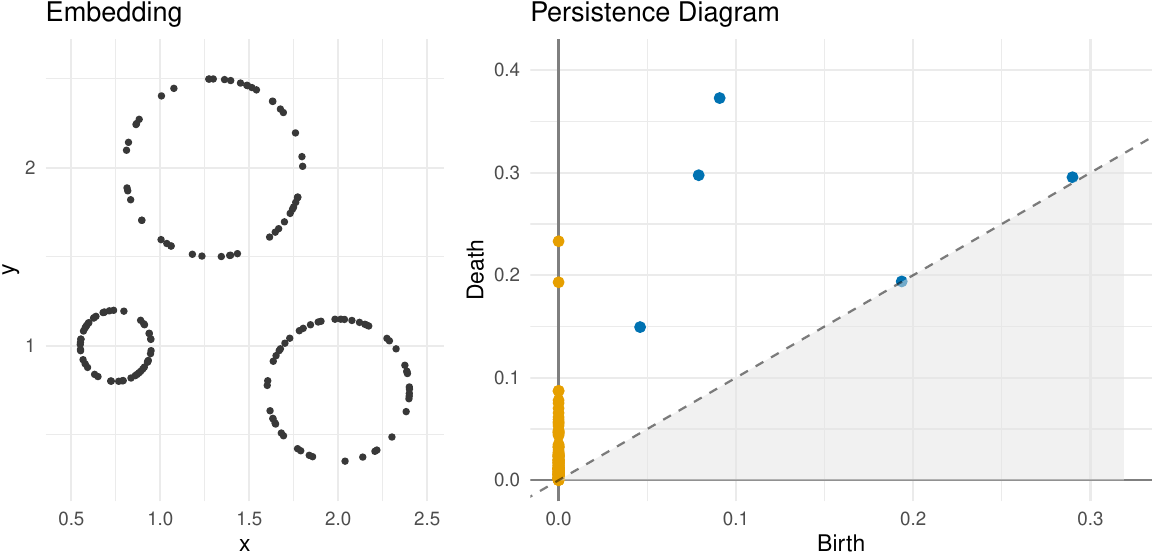} \end{center}
\caption{H0 and H1 persistence diagrams based on an embedding space}
\label{fig:hom-dims}
\end{figure*}

Figure \ref{fig:hom-dims} shows the familiar three ring embedding space
alongside the \(H0\) and \(H1\) persistence diagram. Rather than moving
off the vertical axis like the \(H1\) graph points, the \(H0\) points
lie along the vertical axis. This alignment occurs because each point is
born immediately and dies (following the elder rule) when it connects to
another point or group of points. Each singleton eventually merges with
another group until every point is part of one large feature.

The \(H0\) persistence diagram shows three main phases. First, many
merging events occur as the close points within each ring merge
together. Then, after a gap, one ring connects to another. Finally, the
third ring connects to the other two, creating a single feature
containing all points. Since this fully connected feature cannot merge
with another group, it never dies. This feature is said to have infinite
persistence and is called the ``essential'' part of the persistence
diagram. While constructing the following representations, this point is
typically dropped.

\subsubsection{Rotated and Rescaled Persistence
Diagram}\label{rotated-and-rescaled-persistence-diagram}

\begin{figure*}[!t]
\centering

\begin{center}\includegraphics[width=1.0\textwidth]{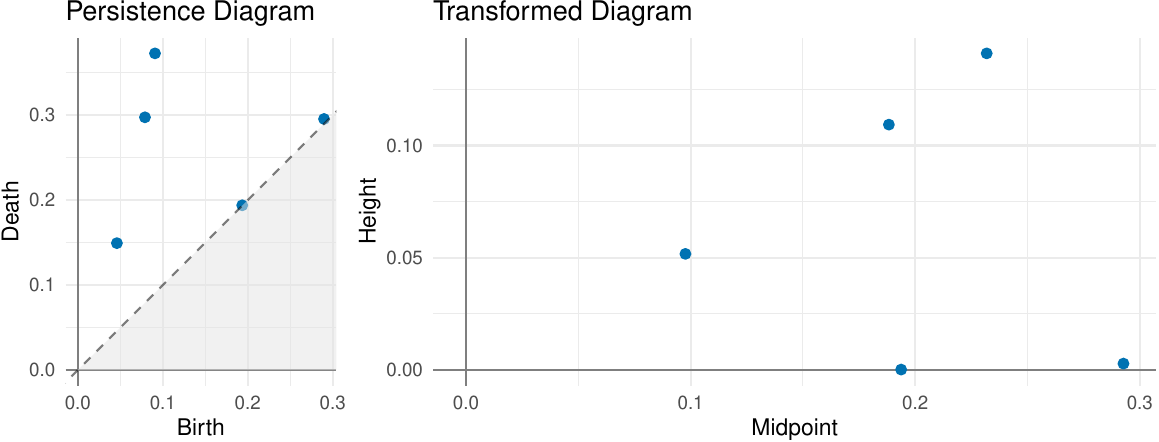} \end{center}
\caption{Transformation of a H1 persistence diagram with standard transformation parameters}
\label{fig:pd-transformation}
\end{figure*}

All points of the persistence diagram lie in the region
\(\left\{(b,d)\in \mathbb{R}^2 | d\geq b, b,d>0 \right\}\), so the
region of Q1 below the diagonal offers no information. Since this area
is uninformative, rotation and rescaling transformations are applied as
seen in Figure \ref{fig:pd-transformation}.

For transition matrix \(\phi: (b,d) \to (m,h)\) and \(b\leq d\),

\begin{singlespace}
\[
\begin{bmatrix} m \\ h \end{bmatrix} = 
    \phi(\theta, \alpha) \begin{bmatrix} b \\ d \end{bmatrix},
\quad
\phi(\theta, \alpha) = \alpha
    \begin{bmatrix}
        cos\theta & -sin\theta \\
        sin\theta & cos\theta
    \end{bmatrix}
\]
\end{singlespace}

Typically \(\alpha = \frac{\sqrt{2}}{2}\), \(\theta=-\frac{\pi}{4}\),
\(m=\frac{b+d}{2}\), and \(h=\frac{b-d}{2}\) for the rotated and
rescaled persistence diagram.

\begin{figure*}[!t]
\centering

\begin{center}\includegraphics[width=1\linewidth]{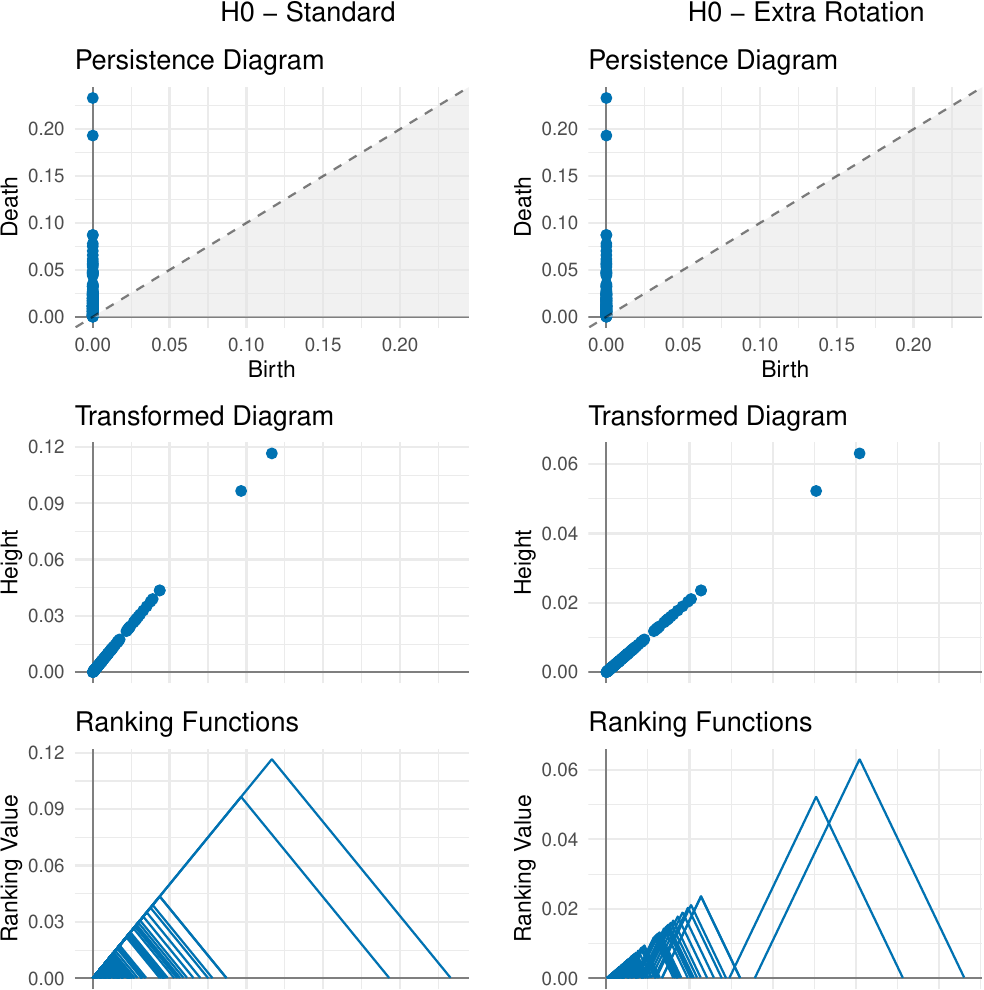} \end{center}
\caption{Demonstration of how standard transformation parameters produce colinear ranking functions for H0, and how the updated transformation parameters fix this issue}
\label{fig:pd-colinearity-h0}
\end{figure*}

Since the \(0^{\mathrm{th}}\) homology dimension (\(p=0\)) persistence
diagram tracks connected components, the initial features are singletons
from the point cloud and are born immediately such that \(b=0\) for all
\(0\)-dimensional persistence pairs. A ranking function is introduced in
the next section, \ref{section:ranking-functions}, but Figure
\ref{fig:pd-colinearity-h0} shows how the standard \(H0\) transformed
persistence diagram results in ranking functions with colinear left
legs. To avoid the colinearity, this work sets
\(\theta = -\frac{3\pi}{8}\) when transforming \(H0\) persistence
diagrams. Since no \(H0\) points exist off the vertical axis and the
transformation is still isometric in \(\mathbb{R}^2\), the
non-negativity, monotonicity, and stability guarantees of the
persistence landscapes hold as introduced below. While this colinearity
creates uninformative persistence landscapes, it is not an issue for
persistence silhouettes (also introduced below), but the extra rotation
is applied for consistency.

\subsection{Persistence Diagrams to Aggregated
Representations}\label{section:ranking-functions}

\begin{figure}[!t]
\centering

\begin{center}\includegraphics[width=1\linewidth]{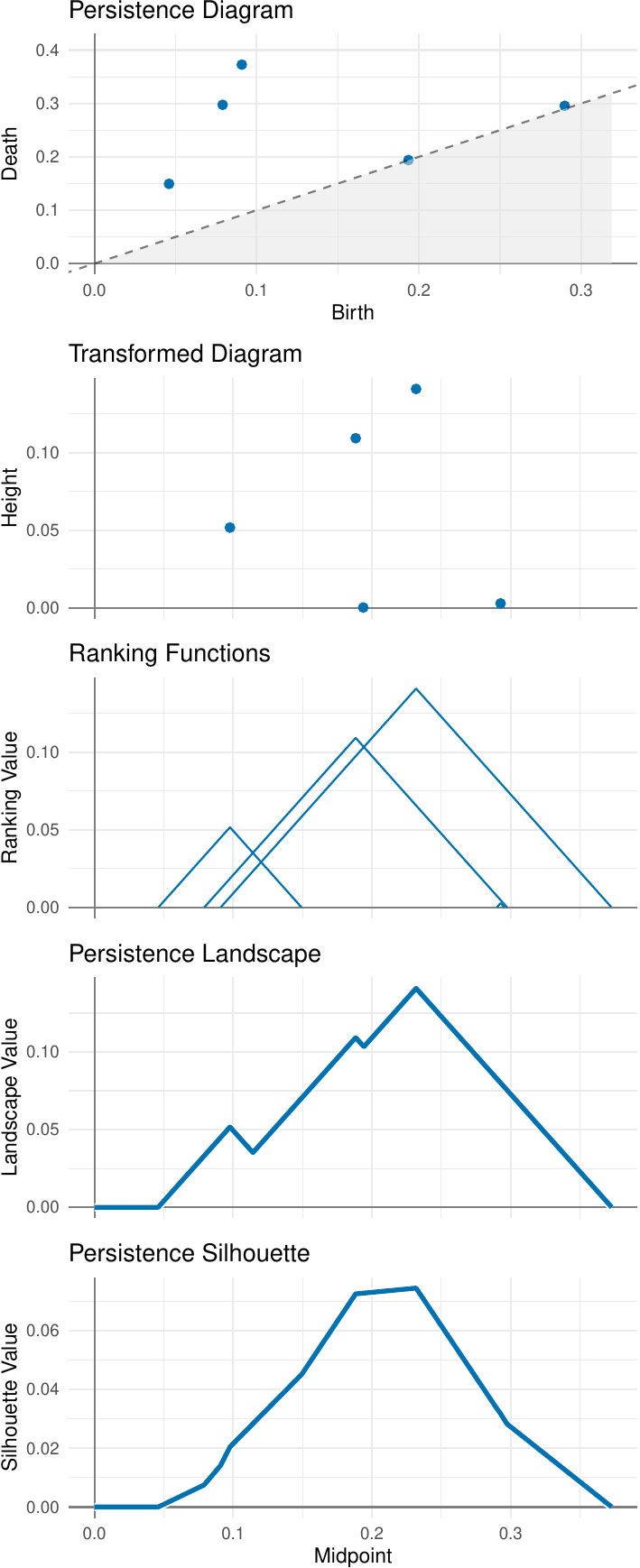} \end{center}
\caption{Full process of producing persistence landscapes and silhouettes from persistence diagrams.}
\label{fig:pd-to-rep}
\end{figure}

\noindent

Once the persistence diagrams are transformed, a ranking function is
defined and aggregation methods are used to construct the landscapes and
silhouettes. Figure \ref{fig:pd-to-rep} shows each step from persistence
diagram to the aggregated representations.

The third graph shows the new ranking functions built around the points
of the transformed persistence diagram. The ranking functions are simply
triangles centered on each point. These are similar to kernels in kernel
density estimation (KDE) but they are not proper probability kernels
since they do not sum to one.

The fourth and fifth graph show the persistence landscape and silhouette
respectively. They both aggregate the ranking functions in different
ways. The landscapes take the maximum value of the ranking function at
each point while the silhouettes sum the ranking functions then
normalize the values to create a probability distribution.

These summary functions do differ in important ways. While the
landscapes capture dominant persistence features at different scales,
the silhouette and KDE approaches aggregate the effect of persistence
across scales. Both methods have benefits, but it is important to note
the change from the maxitive kernels of the landscapes to the summative
kernels of the other approaches and understand the differences in the
resulting representations. For example, while the silhouettes are
closely linked to probability theory where summative kernels are common,
maxitive kernels are more common in possibility theory, which has
connections to multicriteria decision making.

\subsubsection{\texorpdfstring{Persistence Landscapes and Silhouettes
\label{section:land-sil}}{Persistence Landscapes and Silhouettes }}\label{persistence-landscapes-and-silhouettes}

\begin{singlespace}
First define a rank function using standard Betti number notation, $\lambda: \mathbb{R}^2 \to \mathbb{R}$,
\[\lambda(b, d)= 
    \begin{cases} 
        \beta^{b, d}, & b\leq d \\
        0, & otw 
    \end{cases}
\]

Then change coordinates using the transformation matrix $\phi(\theta, \alpha)$ to obtain a rotated and rescaled rank function, $\lambda: \mathbb{R}^2 \to \mathbb{R}$,

\[\lambda(m, h)= 
    \begin{cases} 
        \beta^{m-h, m+h}, & h\geq 0 \\
        0, & otw 
    \end{cases}
\]

\begin{definition}[\cite{bubenik2015statistical}]
\label{def:persistence-landscape-definition}
Persistence landscape are built from the rescaled rank function. For a sequence of functions $\lambda_k: \mathbb{R}\to\overline{\mathbb{R}}$, where $\lambda_k(t) = \lambda(k,t)$, $$\lambda_k(t) = sup\left(m \geq 0 \mid \beta^{t-m, t+m} \geq k\right)$$

and have the following guarantees,
\begin{align*}
    1.\quad & \lambda_k(t)\geq 0, & \text{non-negative} \\
    2.\quad & \lambda_k(t)\geq \lambda_{k+1}, & \text{monotonic decreasing} \\
    3.\quad & \lambda_k \text{ is \textit{1-Lipschitz}}, & \text{stable}
\end{align*}
\end{definition}
\end{singlespace}

Persistence landscapes are a representation of persistence diagrams
without their essential parts and can be computed in
\(O\left(n exp(m)\right)\) \autocite{avg-landscape-stab}.

More intuitively, the persistence landscape can be defined as a set of
continuous, piecewise linear functions
\(\left\{ \Lambda_c\right\}_{c\in D}\) created by ``tenting'' each
point, \(c=(m,h)\), of the rotated and rescaled persistence diagram,
\(D\) \autocite{pers-silh}:

\begin{singlespace}
\[\Lambda_c(t)= 
    \begin{cases} 
        t-m+h, & t \in [m-h, m] \\
        m+h-t, & t \in [m, m+h] \\
        0, & otw 
    \end{cases}
\]

and $$\lambda_k(t) = \underset{c\in D}{kmax}\;\Lambda_c(t), \quad t\in [0,T],\; k\in \mathbb{N}$$

\begin{definition}[\cite{pers-silh}]
\label{def:persistence-silhouette-definition}
A weighted persistence silhouette that takes a weighted average of the ranking functions, $$\phi(t) = \frac{\sum^m_{j=1} w_j \Lambda_j(t)}{\sum^m_{j=1} w_j}$$
\end{definition}
\end{singlespace}

When the weights take into account the persistence of each point,
\(w_j = \left|d_j - b_j \right|^q\quad \forall\; 0<q\leq\infty\), this
is called the ``Power-Weighted Silhouette''. For any choice of
non-negative weights, this representation is still \(1-Lipschitz\). This
work sets \(q=\frac{1}{2}\).

Changing the value of \(q\) highlights different parts of the
persistence diagram. If \(q\) is small, low persistence points are more
prominent, and if \(q\) is large, high persistence points are more
prominent. This is a helpful tuning parameter, and aligns with more
recent representations of persistence diagrams that rely on Kernel
Density Estimation (KDE), like persistence images
\autocite{pers-images}.

\begin{figure*}[p]
\centering

\begin{center}\includegraphics[width=0.96\linewidth]{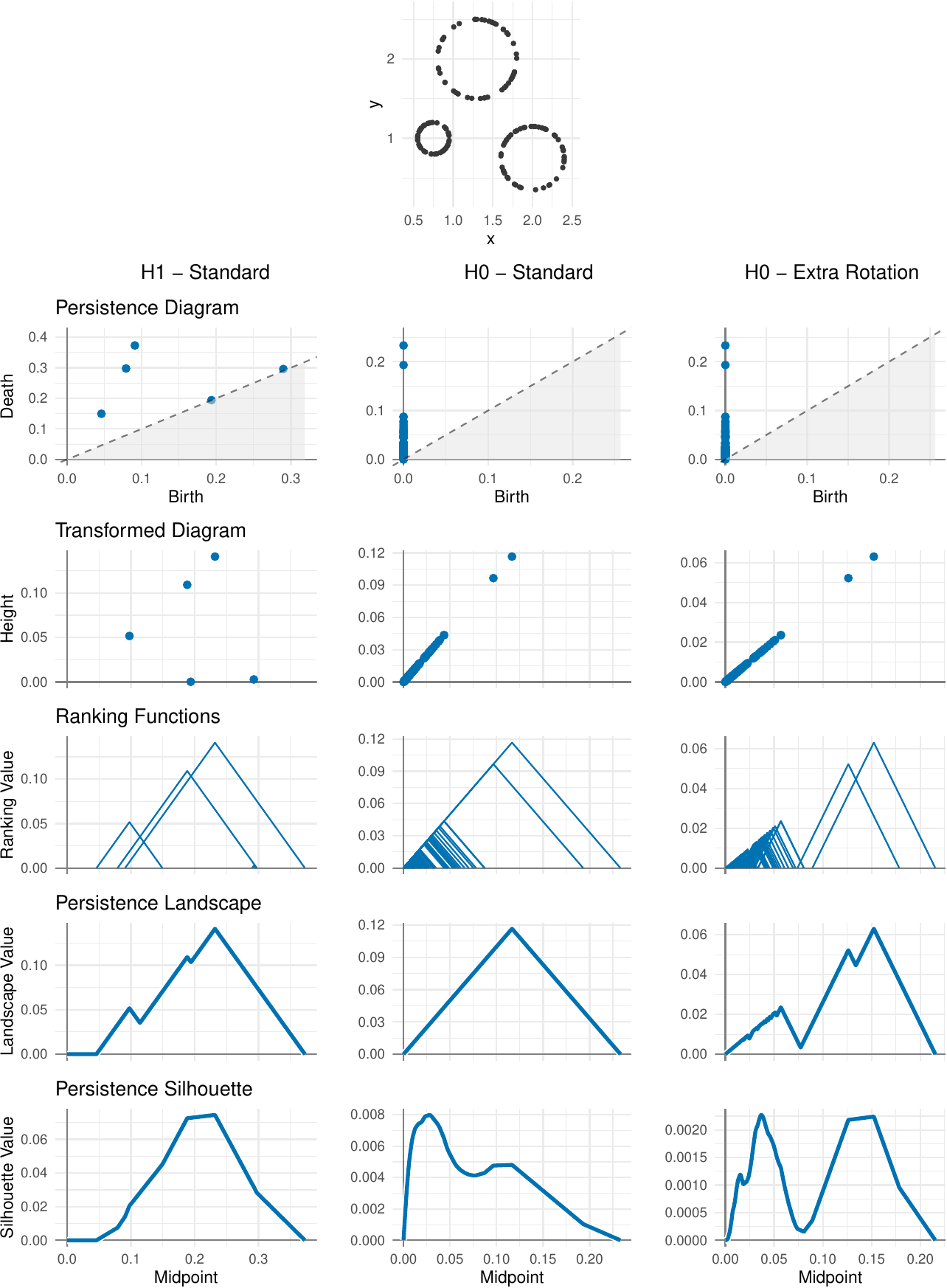} \end{center}
\caption{Landscape and silhouette comparison for a common embedding but differing homology dimensions and transformations}
\label{fig:pd-comparing-dims}
\end{figure*}

Figure \ref{fig:pd-comparing-dims} shows how each stage of the process
changes for an embedding based on the choice of homology group and
applied transformation. It is clear that different features are
highlighted by each technique. The colinearity issue is especially
apparent in the standard \(H0\) method. While the three rings are clear
in the landscape peaks of the \(H1\) and \(H0\) with extra rotation,
this information is completely lost in the standard \(H0\) landscape.

\subsubsection{Mean Representations}\label{mean-representations}

To ensure that the representations are not overfit to a single
realization of an embedding space, mean representations are constructed
by bootstrapping subsamples of the graph and averaging the associated
landscapes or silhouettes.

\begin{figure*}[!t]
\centering

\begin{center}\includegraphics[width=1\linewidth]{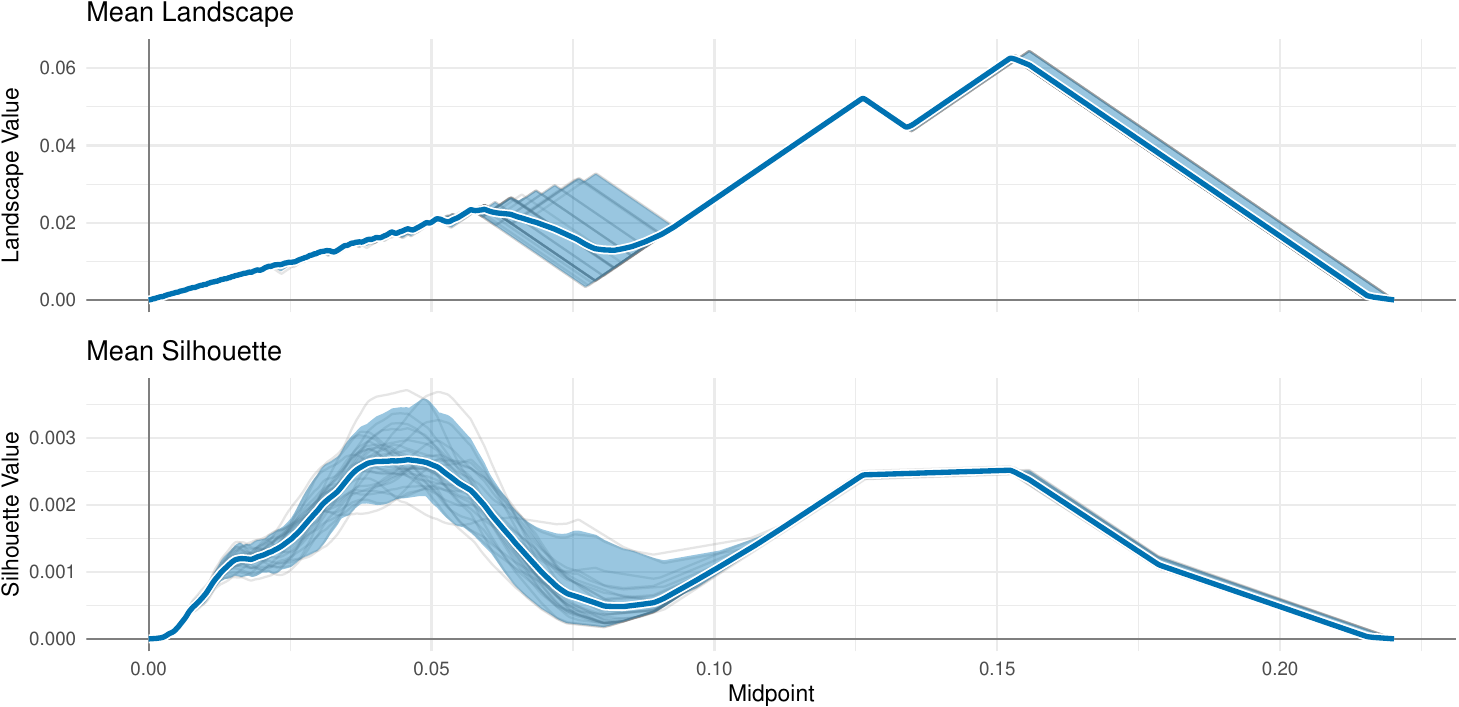} \end{center}
\caption{Mean landscape and silhouettes with point-wise 95\% confidence bounds built from bootstrapped representations} 
\label{fig:mean-reps}
\end{figure*}

Let \(\left(\mathbb{X}, \rho\right)\) be a metric space with a metric,
\(\rho\), diameter, \(d\), \(0\leq d\leq T/2\) and
\(\mu\in \mathcal{P}\left(\mathbb{X}\right)\) where its support,
\(\mathbb{X}_{\mu}\), is a compact set and \(T\) is defined as in
Section \ref{section:land-sil}.

For \(X = \{x_1,..., x_m\}\subset \mathbb{X}\), a sample of \(m>0\)
points from \(\mu \in \mathcal{P}\left(\mathbb{X}\right)\),
\(\lambda_X\) is the persistence landscape, \(\Psi_{\mu}^{m}\) is the
measure induced by \(\mu^{\otimes m}\), and
\(E_{\Psi_{\mu}^m}[\lambda_X]\) is the average landscape.

\begin{singlespace}
\begin{definition}[\cite{avg-landscape-stab}]
\label{def:average-landscape-definition}
For, $S_1^m, ..., S_n^m$, $n$ independent samples of size $m$ from $\mu$, the empirical average landscape that estimates $\lambda_{\mathbb{X}_\mu}$ is,
$$\overline{\lambda_n^m}(t) = \frac{1}{n} \sum^n_{i=1}\lambda_{i}^m(t) \quad \forall t\in [0, T]$$ 
\end{definition}
\end{singlespace}

The mean persistence landscape provides a functional representation to
use in statistical analyses \autocite{tda-intro}.
\textcite{bubenik2015statistical} explore formal homogeneity testing and
\textcite{avg-landscape-stab} use derived confidence bounds to test the
dissimilarity of point clouds based on \(\lambda_\infty\) distances over
the average landscape.

The \(95\%\) confidence interval is derived point-wise using a standard
z-score based on the bootstrapped samples. These bands are used
primarily for intuition about the sample variance in creating the mean
representations. Other work has outlined confidence bands for landscapes
and silhouettes, but they are either designed for bounds on the norm of
the representation (\cite{bubenik2015statistical}) or have convergence
rates that are prohibitively slow (\cite{stab-pm}). The choice of
displayed confidence bound does not affect the testing in this work, a
non-parametric method that does not rely on z-scores or strict
distributional assumptions is used for the formal testing.

\subsection{Scoring Representation
Similarity}\label{scoring-representation-similarity}

\begin{figure*}[p]
\centering

\begin{center}\includegraphics[width=1\linewidth]{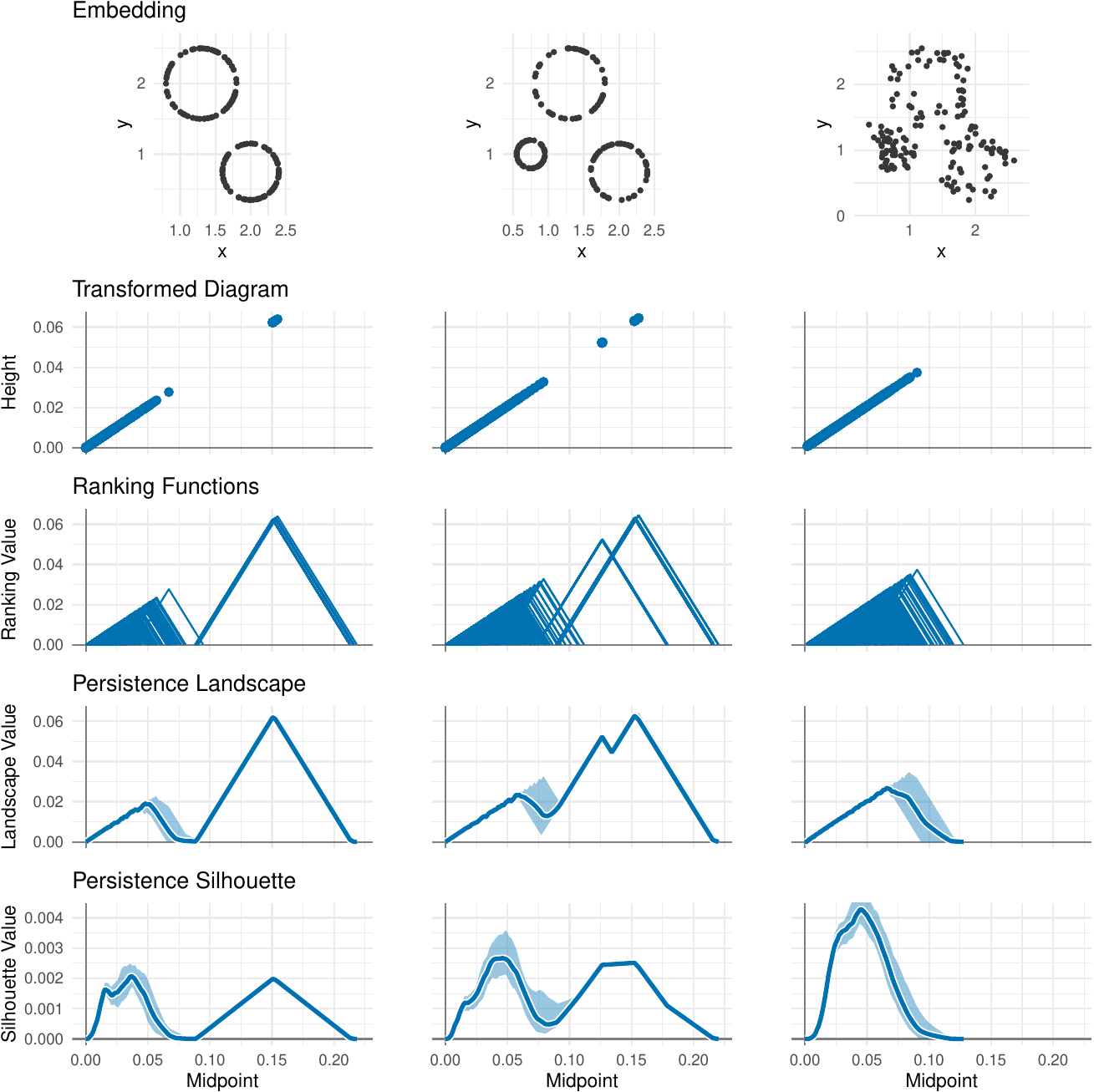} \end{center}
\caption{Comparing H0 landscape and silhouette representations across three different embeddings}
\label{fig:pd-comparing-embeddings}
\end{figure*}

Figure \ref{fig:pd-comparing-embeddings} shows how the mean
representations change as the embedding changes. On the left and the
middle, the clean rings show up as mean landscape peaks. The right
column, however, shows that as noise is added to the rings, the
structural coherence of the rings breaks down and the mean
representations shift dramatically. Rigorous two-sample homogeneity
tests can be built around energy statistics to test these changes.

\subsubsection{Energy Statistics}\label{energy-statistics}

Energy statistics based methods \autocite{you2022comparing} can compare
landscapes generated using bootstrapping \autocite{avg-landscape-stab}.

\begin{singlespace}
\begin{definition}[\cite{you2022comparing}]
\label{def:energy-distance-definition}
Given independent random samples $X_1, ..., X_m$ and $Y_1, ..., Y_n$, the empirical energy distance between $X$ and $Y$ is, 

$$\begin{aligned} \mathcal{E}_{m,n}(X, Y) &= \frac{2}{mn} \sum^m_{i=1}\sum^n_{j=1} \left\lVert X_i - Y_j \right\rVert \\ &- \frac{1}{m^2} \sum^m_{i=1}\sum^m_{i'=1} \left\lVert X_i - X_{i'} \right\rVert \\ &- \frac{1}{n^2} \sum^n_{j=1}\sum^n_{j'=1} \left\lVert Y_i - Y_{j'} \right\rVert \end{aligned}$$
\end{definition}
\end{singlespace}

The distance combines a measurement of the difference of samples between
embeddings X and Y and of the differences between samples within
embeddings X and Y.

From the energy distance, the scaled energy test statistic is,
\[T_{m,n} = \frac{mn}{m+n} \mathcal{E}_{m,n}\left(X,Y\right)\]

\subsubsection{Formal Testing}\label{formal-testing}

\textcite{equal-dist-high-dim} show that this is a consistent
homogeneity test against all fixed hypotheses. If all possible
permutations of the sampled data are used, the test is exact, but since
this is often computationally infeasible, bootstrap methods are
recommended.

Extending to landscapes, let
\(\pmb\lambda_1 = \lambda_{1,1}, ..., \lambda_{1,n_1}\) and
\(\pmb\lambda_2 = \lambda_{2,1}, ..., \lambda_{2,n_2}\) be independent
random samples from the Borel probability measures \(\mu_1\), \(\mu_2\).
These samples are averaged to create the mean representation. Then the
two sample test statistic is,
\[\begin{aligned} T_{n_1, n_2}\left(\pmb\lambda_1, \pmb\lambda_2\right) = \frac{n_1 n_2}{n_1+n_2}\Bigg[ \frac{2}{n_1 n_2}\sum^{n_1}_{i=1}\sum^{n_2}_{j=1} \delta_2\left(\lambda_{1,i}, \lambda_{2, j}\right) \\ - \frac{1}{n_1^2}\sum^{n_1}_{i,i'=1}\delta_2\left(\lambda_{1,i}, \lambda_{1,i'}\right) \\ - \frac{1}{n_2^2}\sum^{n_2}_{j,j'=1}\delta_2\left(\lambda_{2,j}, \lambda_{2,j'}\right) \Bigg] \end{aligned}\]
where
\[\begin{aligned} \delta_2\left(\lambda, \lambda'\right) &\coloneqq \left\lVert \lambda - \lambda' \right\rVert_2 \\ &= \left(\sum^{\infty}_{k=1}\left[\int_{-\infty}^{\infty} \left( \lambda_k(t) - \lambda'_k(t)\right)^2 dt\right] \right)^{1/2} \end{aligned}\]
The non-parametric p-value is computed from a set of permuted test
statistics, \(\left\{T^{(b)}_{n_1, n_2}\right\}_{b=1}^B\)
\autocite{you2022comparing}.

For the pooled sample
\(\left\{\tilde\lambda_i\right\}^{n_1+n_2}_{i=1} = \pmb\lambda_1 \bigcup \pmb\lambda_2\)
and \(\pmb\sigma = (\sigma(1), ..., \sigma(n))\), a random vector
uniformly distributed over all permutations of \(\{1,...,n\}\) where
\(\left\{\pmb\sigma_b\right\}_1^B\) are i.i.d. copies of \(\pmb\sigma\),
the permuted landscape sets are,
\[\pmb\lambda_1^{(b)} = \tilde\lambda_{\pmb\sigma_b(1)}, ..., \tilde\lambda_{\pmb\sigma_b(n_1)}, \quad \pmb\lambda_2^{(b)} = \tilde\lambda_{\pmb\sigma_b(n_1+1)}, ..., \tilde\lambda_{\pmb\sigma_b(n_1+n_2)}\]
Each \(T^{(b)}_{n_1, n_2}\) is based on permuted landscape sets
\(\pmb\lambda_1^{(b)}\), \(\pmb\lambda_2^{(b)}\) and the p-value is,
\[\hat{p} = \frac{1}{B+1}\left( \sum_{b=1}^B I\left(T_{n_1, n_2} \leq T_{n_1,n_2}^{(b)}\right) + 1 \right)\]
Together, this results in a test statistic and a homogeneity test at a
desired \(\alpha\)-significance level for \(\mu_1\) and \(\mu_2\) from
which \(\pmb\lambda_1\) and \(\pmb\lambda_2\) are sampled,
\[H_0: \mu_1 = \mu_2, \quad H_a: \mu_1 \neq \mu_2\]

Although they are defined using the landscapes \(\lambda\), these tests
can also be done with the persistence silhouettes by replacing
\(\lambda\) with \(\phi\) in the process above.

To control the false discovery rate, when making multiple comparisons in
the results, the Benjamini--Yekutieli (\cite{by-p-value}) procedure is
used to adjust the significance threshold within the landscape or
silhouette groups.

The energy statistics results are also compared to methods based on
curve matching. In each result two additional metrics are included,
where the p-values for these metrics are computed using the same
permutation test as the energy statistics: the Wasserstein Distance,
which is a natural metric for landscapes
\[W(P,Q) =  \underset{J\in \mathcal{J}(P,Q)}{\inf} \int_{\mathbb{R} \times \mathbb{R}} \left| x-y \right| dJ(x,y)\]
and Jensen-Shannon Distance (JSD) which is a natural metric for
silhouettes.

\begin{singlespace}
$$\begin{aligned} JSD(P, Q) &= \sqrt{D_{JS}(P\parallel Q)} \\ &= \sqrt{\frac{D_{KL}(P\parallel M) + D_{KL}(Q\parallel M)}{2}} \end{aligned}$$
where
\begin{align*}
    M &= \frac{P+Q}{2} \\
    D_{KL}(P \parallel Q) &= \sum_{x\in \mathbb{X}} P(x) log\left(\frac{P(x)}{Q(x)}\right)
\end{align*}
\end{singlespace}

\begin{figure*}[p]
\centering

\begin{center}\includegraphics[width=0.86\linewidth]{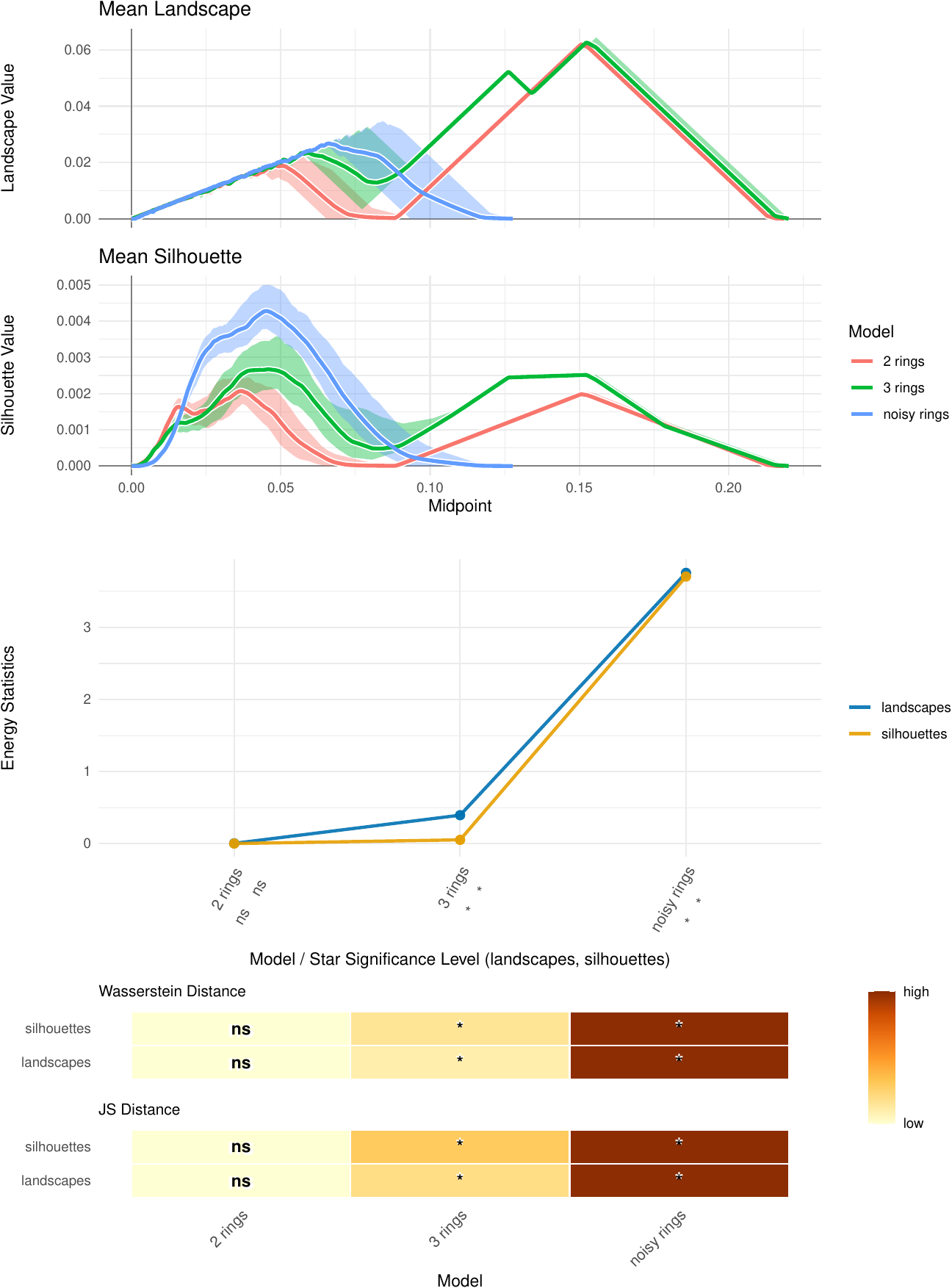} \end{center}
\caption{Ring Example Results: (top) Mean representations of three embeddings with 95\% point-wise confidence bounds. (middle) Line plot of test statistics comparing each embedding to the two ring baseline with p-values represented in star notation under the model labels. (bottom) Heatmaps with the Wasserstein distance and JS distance between three embeddings and the two ring baseline.}
\label{fig:energy-stat-heatmap-line-example}
\end{figure*}

Figure \ref{fig:energy-stat-heatmap-line-example} shows the results of
comparing the three embedding spaces from Figure
\ref{fig:pd-comparing-embeddings}. The top plots show the mean
representations with the associated point-wise confidence bands. The
middle plot shows the energy statistics of each representation compared
to the two ring embedding as the baseline along with the associated
p-values in star-notation below the model name. On the bottom, the
Wasserstein and JS distances are shown as heatmaps where the values are
normalized within each representation. The energy statistics are the
primary metric, but the heatmaps are presented as a guide to how
different metrics capture the space. Generally the results show
directional alignment between the three metrics, but especially in the
section with less tightly controlled models, the Wasserstein and JS
distances prove to be less discriminative than the energy statistics.

From the graphs it is easy to see that the embeddings with clear rings
appear to be more similar than to the noisy embedding space. In the line
plot and the heatmaps, each embedding is compared to the two ring
embeddings. The two ring model has a score of 0 and non-significant
p-values, as it should be since it is being compared to itself. The
tests detect significant differences between the two ring baseline and
the three ring and noisy embeddings. The tests indicate that the two
ring embedding is more similar to the three ring embedding than the
noisy embedding. This is an important example because it reinforces that
the tests are picking up structural coherence and differentiate between
clean and noisy representations.

\subsection{Stability}\label{stability}

In all of the energy statistics plots, each dot represents a comparison
between two models' embedding of the same input data. These test
statistics can change if either the models change or the underlying data
changes. This work focuses on changing the models while holding the data
constant, but leaves open the question of how much variability new
datasets introduce into the energy statistics.

However, the bootstrapped representations which make up the energy
statistics are already developed from subsamples of the data. A
substantially new dataset would need to be created or added into the
existing dataset if variability is to be tested well. This resilience to
small data changes is a strength of the Lipschitz stable representations
and the non-parametric energy statistics, but it makes it more difficult
to correctly characterize their true sensitivity.

Finding representative data which is sampled from the same underlying
distribution would also be difficult. If the data is not sampled from
the same distribution, it may be difficult to differentiate between the
variability introduced by the model or by domain shift. The problem may
be easily solved in toy examples where the data distribution is known
but is a significant challenge for real-world data.

\section{Methods}\label{methods}

The approach in this work consists of four phases:

\begin{enumerate}
\def\labelenumi{\arabic{enumi}.}
\tightlist
\item
  Define a data set with archetypes and blurry examples
\item
  Generate embedding spaces through human curation or model extraction
\item
  Create topological representations of the normalized semantic spaces
\item
  Score and test alignment between representations
\end{enumerate}

It is important to remember that these tools do not test alignment
against some absolute truth. The tests are comparative and defining how
concepts relate to each other is difficult to make definitive. This work
assumes only that the relationships have been defined through some
deliberative and auditable process. Whether these representations are
human curated knowledge graphs or machine generated embedding spaces,
the comparisons will be meaningful only if the relationships between the
inputs carry meaning.

For the datasets in this work, defining the relationships means first
identifying core concepts, or archetypes, across multiple domains. These
archetypes represent concepts that should be relatively distinct
semantically. Then, examples from these archetypes are selected that
represent clean examples of a concept or blur the boundaries between
concepts. By capturing how different representations of the datasets are
structured, topological tools give insights to the local coherence of
individual concepts, global relationships between concepts, and
alignment between different representations. The data generation and
embedding methods change for each result based on domain and model
selection. These baselines will be described more in the relevant
Results sections.

\clearpage

\begin{center}\rule{0.5\linewidth}{0.5pt}\end{center}

\section{Results}\label{results}

The framework and tests described above are applied to two image data
sets to demonstrate how abstract, high dimensional embeddings can be
compared to low dimensional, interpretable baselines. The results
demonstrate how model selection and fine-tuning can be guided as
increasingly strong model adaptations are applied.

\subsection{\texorpdfstring{Art Classification Results
\label{section:art}}{Art Classification Results }}\label{art-classification-results}

The first results are produced by training an image classifier to
predict the artist behind an artwork. Art is well suited to this work
because each artist has a unique style, but each body of work contains
variations. By training a model to differentiate between artistic
styles, the model should begin to map the artworks into increasingly
coherent conceptual spaces.

\begin{figure*}[!t]
\centering

\begin{center}\includegraphics[width=1\linewidth]{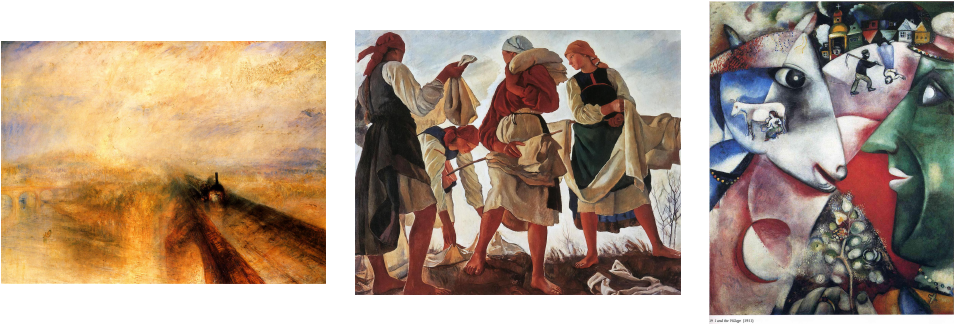} \end{center}
\caption{Example artwork. (left to right) J.M.W. Turner, \textrm{Rain, Steam and Speed - The Great Western Railway}, 1844. Oil on canvas, 91 cm x 121.8 cm. The National Gallery, London; Zinaida Serebriakova, \textrm{Whitening Canvas (also known as Bleaching the Cloth)}, 1917. Oil on canvas, 141.8 cm x 173.6 cm. State Tretyakov Gallery, Moscow; Marc Chagall, \textrm{I and the Village}, 1911. Oil on canvas, 192.1 cm x 151.4 cm. Museum of Modern Art, New York.}
\label{fig:artworks}
\end{figure*}

For the dataset, artworks by Turner, Serebriakova, and Chagall are
pulled from the WikiArt dataset on
Huggingface.\footnote{huggingface.co/datasets/huggan/wikiart} Figure
\ref{fig:artworks} shows the distinctive styles and subject matter of
these three artists, which should help the model classify the work
effectively and allows for the development of a clear baseline
consisting of three disjoint groups.

The base model is \texttt{clip-vit-base-patch32} developed by OpenAI and
available on
Huggingface.\footnote{huggingface.co/openai/clip-vit-base-patch32}
Contrastive Language-Image Pre-training (CLIP) (\cite{clip}) is a
multimodal vision and language model with an image encoder and text
decoder transformer architecture. Developed in 2021, it is no longer
state-of-the-art (SOTA), but it is considered a classic model and is
performant enough for this classification task. Despite its age, CLIP is
used here because it has many architectural similarities with SOTA
models and it is well-studied. Each 768-dimensional artwork embedding is
pulled from the pooled output of the vision encoder.

Instead of having to retrain the full CLIP model, the architecture
enables application of Low-Rank Adaptations (LoRAs) (\cite{lora22}).
LoRAs allow you to adjust the behavior of high-dimensional models by
fine-tuning with low computational cost and without full retraining.

For a pre-trained weight matrix, \(W_0 \in \mathbb{R}^{d\times k}\) and
an update during adaptation \(\Delta W\), the result of a modified
forward pass \(h\) for input \(x\) is, \[h = W_0x + \Delta W x\] For a
LoRA, we set \(\Delta W = BA\) where \(B\in \mathbb{R}^{d\times r}\),
\(A\in \mathbb{R}^{r\times k}\) and \(r \ll min(d, k)\) so the new LoRA
modified forward pass is, \[h = W_0x + BA x\] Training \(B\) and \(A\)
is much less expensive than training \(\Delta W\).

LoRAs have a scaling parameter \(\alpha\) that affects how strong of an
adaptation is applied. The overall effect is scaled by
\(\frac{\alpha}{r}\). For these experiments \(r=8\) and
\[\alpha\in \{1,2,4,8,16,32,64\}\] and also include the base model
without adaptation. These LoRAs are applied to the CLIP model and target
4 attention projections: Query, Value, Key, and Output.

\begin{singlespace}
\begin{equation*}
\text{Typical scaling ratios in practice: } \left\{
\begin{aligned}
  \alpha / r &< 1      &\; \text{subtle} \\
  &= 1      &\; \text{baseline} \\
  &\sim 2   &\; \text{standard} \\
  &> 4      &\; \text{strong}
\end{aligned}
\right.
\end{equation*}
\end{singlespace}

Each model is trained using a \(70/15/15\) train/validation/test split
of \(423\) artworks with balanced classes across artists and a batch
size of \(32\). They train for \(15\) epochs using cross-entropy loss,
\(0.1\) dropout, \(0.001\) learning rate, \(0.0001\) weight decay, and
an Adam optimizer.

\begin{figure}[!t]
\centering

\begin{center}\includegraphics[width=1\linewidth]{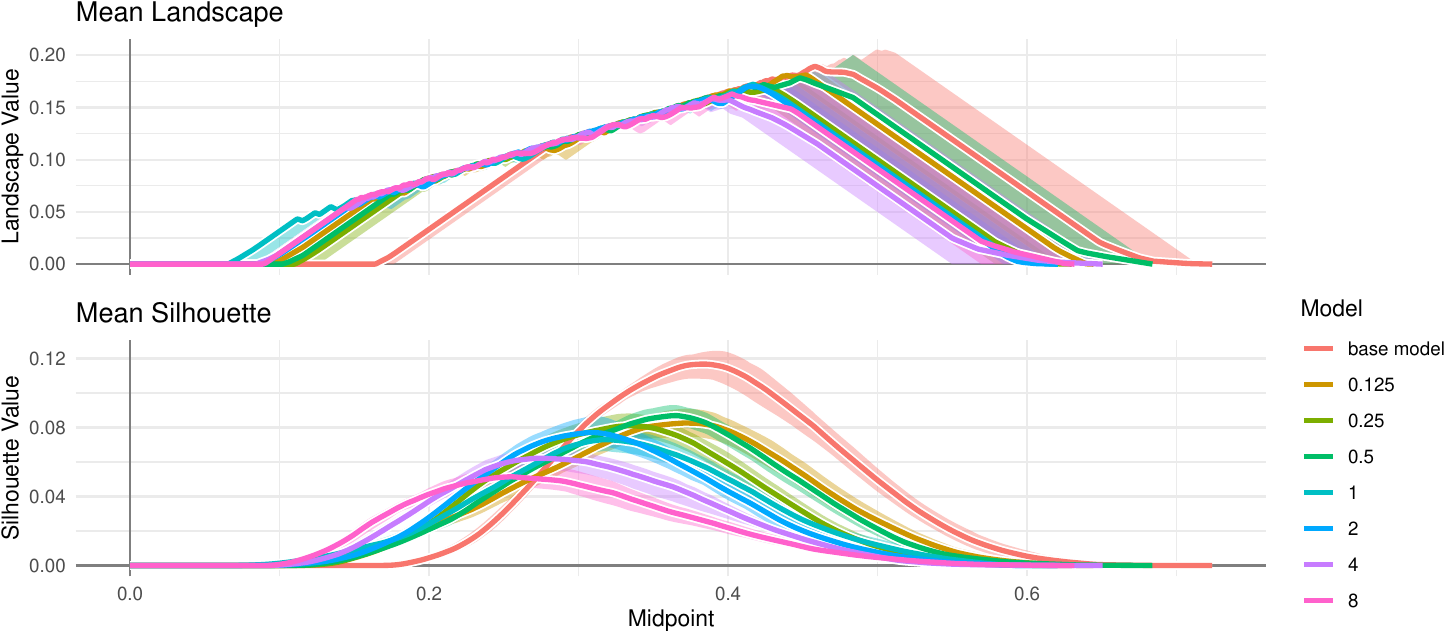} \end{center}
\caption{Mean representations developed from the embedded test set}
\label{fig:art-means}
\end{figure}

Figure \ref{fig:art-means} shows the mean representations along with the
point-wise confidence bands of each of the models. The differences in
these representations are clear from visual inspection, and the formal
tests in the following graphs show the scores reinforce this intuition.

\begin{figure}[!t]
\centering

\begin{center}\includegraphics[width=1\linewidth]{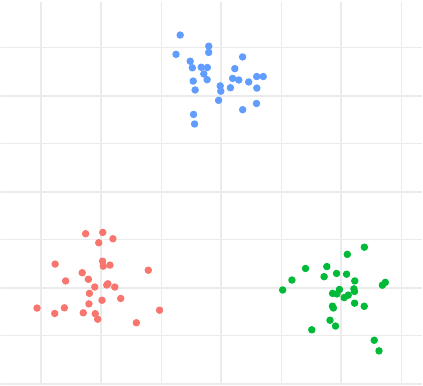} \end{center}
\caption{Curated encoding of the artworks}
\label{fig:art-baseline}
\end{figure}

\begin{figure*}[!t]
\centering

\begin{center}\includegraphics[width=1\linewidth,height=0.45\textheight]{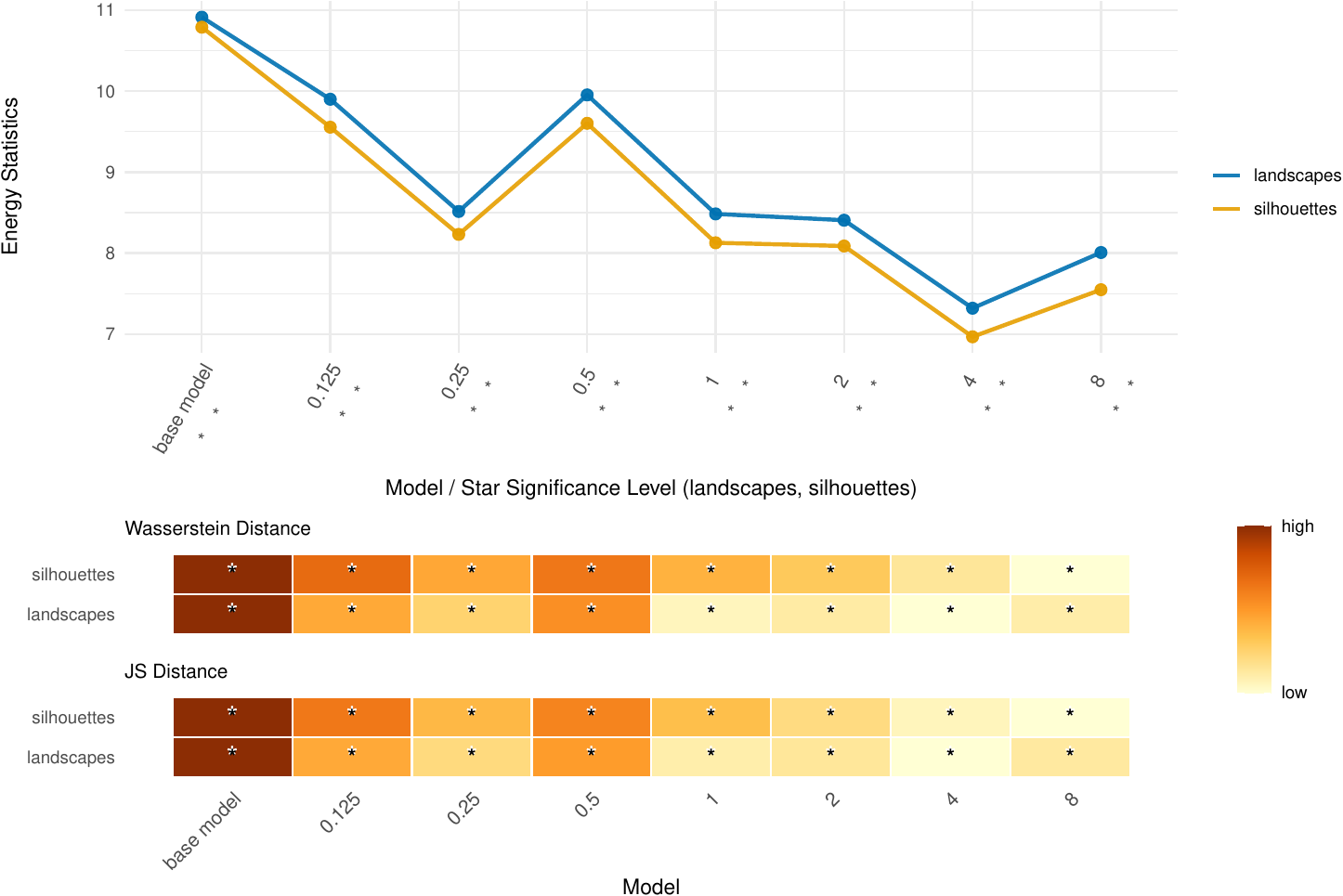} \end{center}
\caption{Comparisons of LoRA fine-tuned model embedding spaces to a curated knowledge encoding}
\label{fig:art-pred-ke}
\end{figure*}

Figure \ref{fig:art-pred-ke} shows the results of applying the LoRAs and
capturing the embeddings of the test set. The energy statistics line
plot and the heatmaps show the comparison of each model to the baseline
displayed in the point cloud in Figure \ref{fig:art-baseline}. Three
distinct groups of points represent the three artists' distinct styles
and the conceptual relationship the model should be learning. This
intuitive baseline makes the plots interpretable, as increasingly
aggressive adaptations are applied to the model, the structure of the
embedding space is increasingly similar to the human defined baseline.

Importantly, Table \ref{tab:art-pred} shows that all of these models are
extremely good at classifying the artworks when judged by outcome
reasoning. The changes in accuracy represent one or two errors with this
relatively small dataset. The test set is nearly saturated, but this
topological technique makes justified model selection possible.

\begin{singlespace}
\begin{table}[H]
\centering
\begin{table}[H]
\centering
\begin{tabular}{cc}
\toprule
$\alpha/r$ & Test Accuracy (\%)\\
\midrule
\cellcolor{gray!10}{0.125} & \cellcolor{gray!10}{98.44}\\
\addlinespace\addlinespace
0.25 & 98.44\\
\addlinespace\addlinespace
\cellcolor{gray!10}{0.5} & \cellcolor{gray!10}{100}\\
\addlinespace\addlinespace
1 & 100\\
\addlinespace\addlinespace
\cellcolor{gray!10}{2} & \cellcolor{gray!10}{100}\\
\addlinespace\addlinespace
4 & 96.88\\
\addlinespace\addlinespace
\cellcolor{gray!10}{8} & \cellcolor{gray!10}{96.88}\\
\bottomrule
\end{tabular}
\end{table}
\caption{Test accuracy across LoRA models}
\label{tab:art-pred}
\end{table}
\end{singlespace}

\begin{figure*}[p]
\centering

\begin{center}\includegraphics[width=0.97\linewidth]{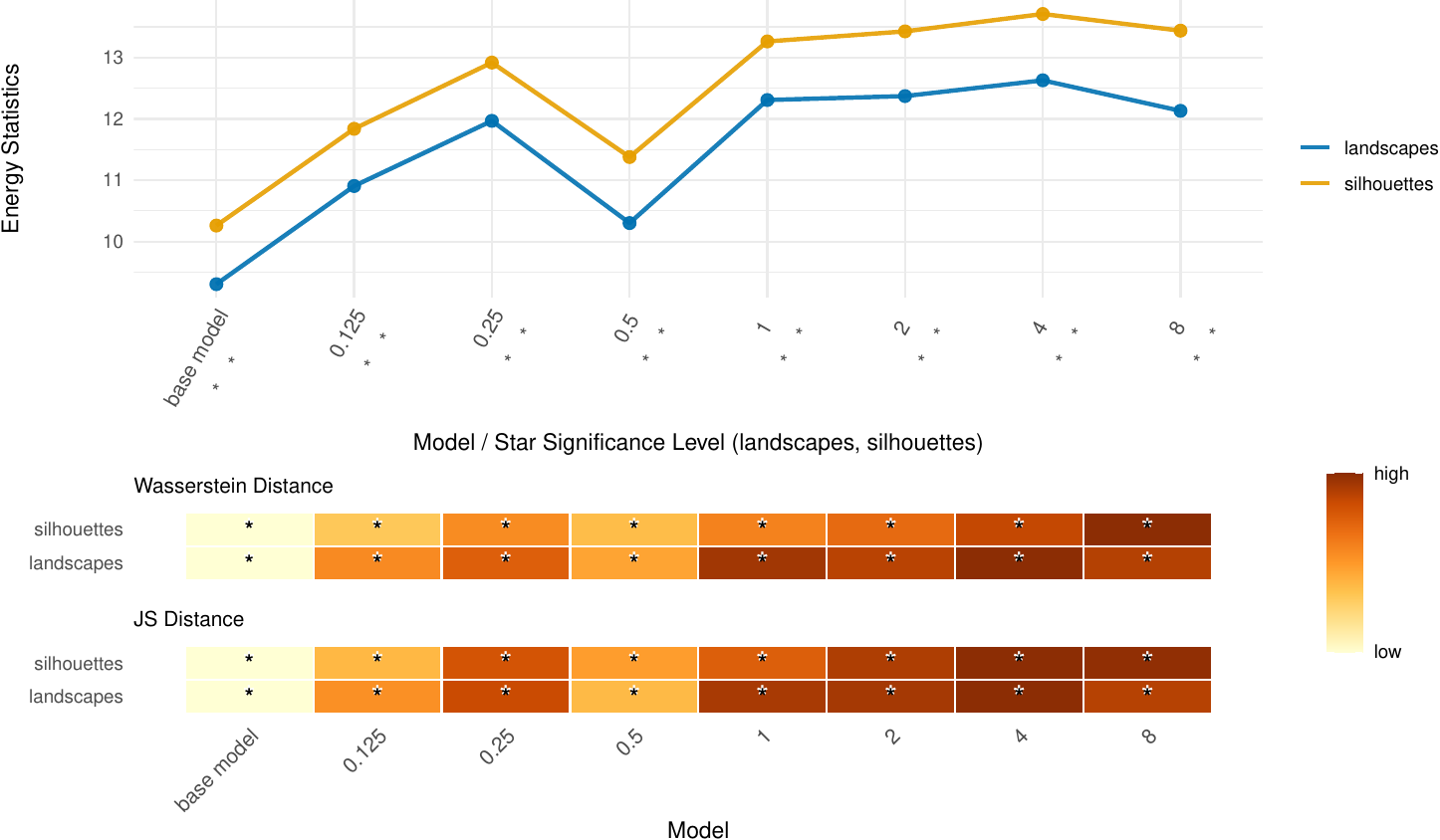} \end{center}
\caption{Comparisons of LoRA fine-tuned model embedding spaces to an equidistant embedding}
\label{fig:art-pred-equi}
\end{figure*}

Figure \ref{fig:art-pred-equi} shows a similar set of charts as Figure
\ref{fig:art-pred-ke}, except the baseline has changed. Instead of an
embedding with three groups, each point is represented as a unit vector
orthogonal to every other point, similar to one-hot encoding where each
data point is a category. This results in an equidistant embedding space
which lacks any broader structural coherence, or alternatively, a space
where each data point is treated as its own distinct archetype. The
energy statistics show a clear trend away from the unstructured
embedding space. As increasingly aggressive LoRAs are applied, the
imposition of structure on the embedding space can be tracked even as
the results on the test set remain similar.

Interestingly, the \(\frac{\alpha}{r}=0.5\) model stands out in both
graphs. This is a perfect example for how a decision between models can
be made. The \(0.5\) and \(1.0\) models both have \(100\%\) accuracy on
the test set, but if a less structured semantic space is preferred the
\(0.5\) model may be preferred over the \(1.0\) model.

\subsection{Model Variation}\label{model-variation}

In Section \ref{section:art}, the \(\frac{\alpha}{r}=0.5\) model stands
out as a sharp deviation from the relatively smooth curve of the other
model adaptations. It is not immediately clear whether this deviation is
due to stochasticity in the model development or if something about this
range of adaptations caused a different embedding structure. New values,
\(\frac{\alpha}{r} \in \{0.375, 0.625, 0.75, 0.875 \}\), are added to
examine what is happening in the region
\(\frac{\alpha}{r} \in [0.25, 1]\).

\begin{figure*}[p]
\centering

\begin{center}\includegraphics[width=0.97\linewidth]{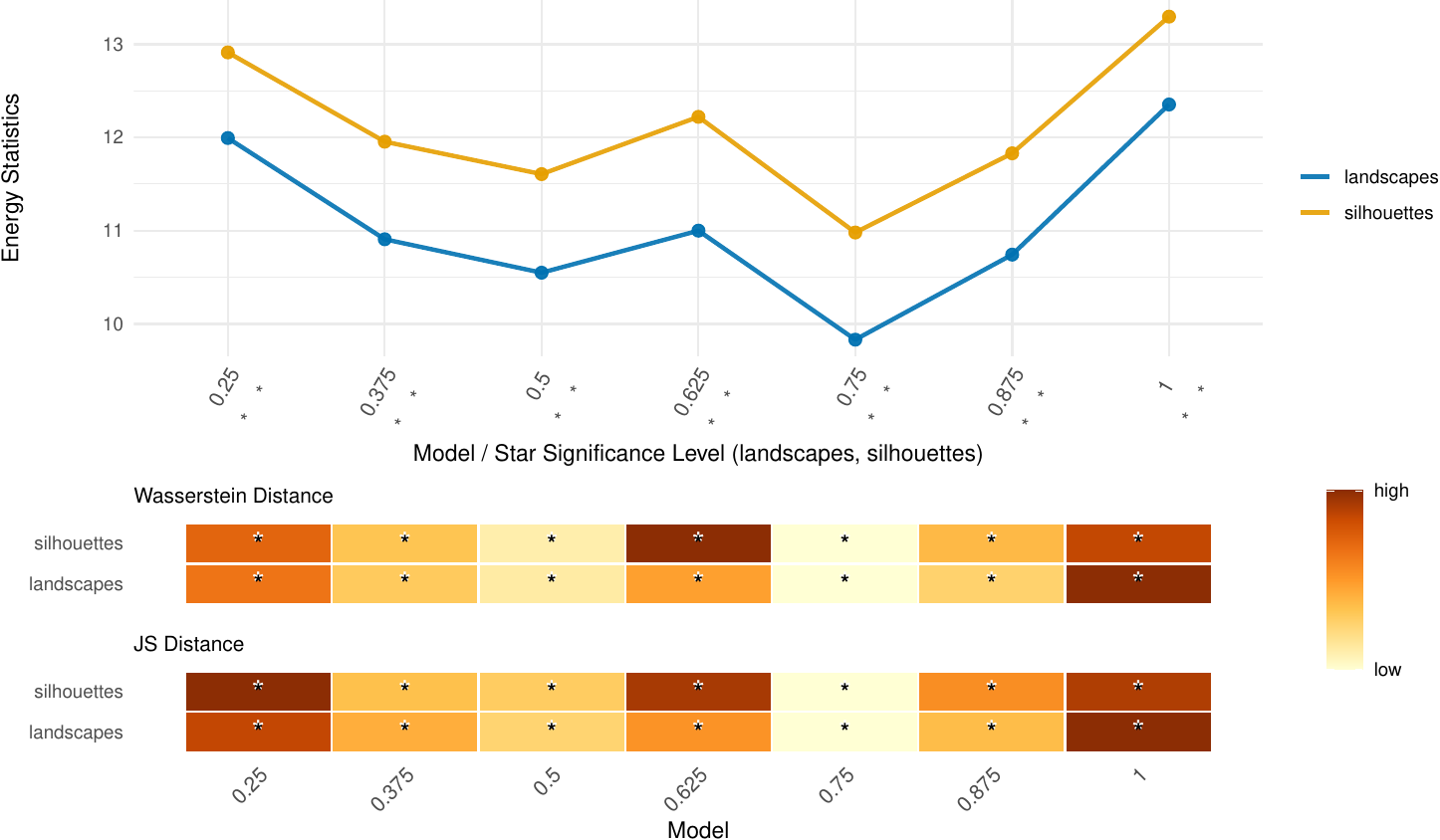} \end{center}
\caption{Comparisons of LoRA fine-tuned model embedding spaces to an equidistant embedding}
\label{fig:art-pred-equi-highlight}
\end{figure*}

Figure \ref{fig:art-pred-equi-highlight} shows this sampled region.
Rather than the \(\frac{\alpha}{r}=0.5\) model being an isolated shift,
it appears that a range of the adaptation are learning a different
semantic structure. Table \ref{tab:art-pred-highlight} shows that the
models are still reaching near perfect test accuracy scores just like
the other adaptations. Since all the training parameters except the
\(\frac{\alpha}{r}\) value were the same across all adaptations, it is
not clear why the structure would change in this region. These
adaptations may be converging on a local minimum in the optimization
surface that is effective for artwork classification but is different
from the minimum found by more aggressive model adaptations.

More research is needed to understand these dynamics. However,
comparisons to different baselines could reveal what kind of structure
this range of adaptations is creating.

\begin{singlespace}
\begin{table}[H]
\centering
\begin{table}[H]
\centering
\begin{tabular}{cc}
\toprule
$\alpha/r$ & Test Accuracy (\%)\\
\midrule
\cellcolor{gray!10}{0.25} & \cellcolor{gray!10}{98.44}\\
\addlinespace\addlinespace
0.375 & 100\\
\addlinespace\addlinespace
\cellcolor{gray!10}{0.5} & \cellcolor{gray!10}{100}\\
\addlinespace\addlinespace
0.625 & 98.44\\
\addlinespace\addlinespace
\cellcolor{gray!10}{0.75} & \cellcolor{gray!10}{100}\\
\addlinespace\addlinespace
0.875 & 100\\
\addlinespace\addlinespace
\cellcolor{gray!10}{1} & \cellcolor{gray!10}{100}\\
\bottomrule
\end{tabular}
\end{table}
\caption{Test accuracy across LoRA models}
\label{tab:art-pred-highlight}
\end{table}
\end{singlespace}

\subsection{Polygonal Results}\label{polygonal-results}

This experiment focuses on a dataset that has clear transitions between
archetypes. While the artwork dataset tried to defined three distinct
clusters, this dataset identifies six archetypes and defines smooth
transitions between them so that most of the data exists in the blurry
space between the rigid concepts. Artistic styles are also not rigidly
defined, but in this dataset, the blurring is directly controlled.

\begin{figure}[!t]
\centering
\centering
\includegraphics[width=\linewidth]{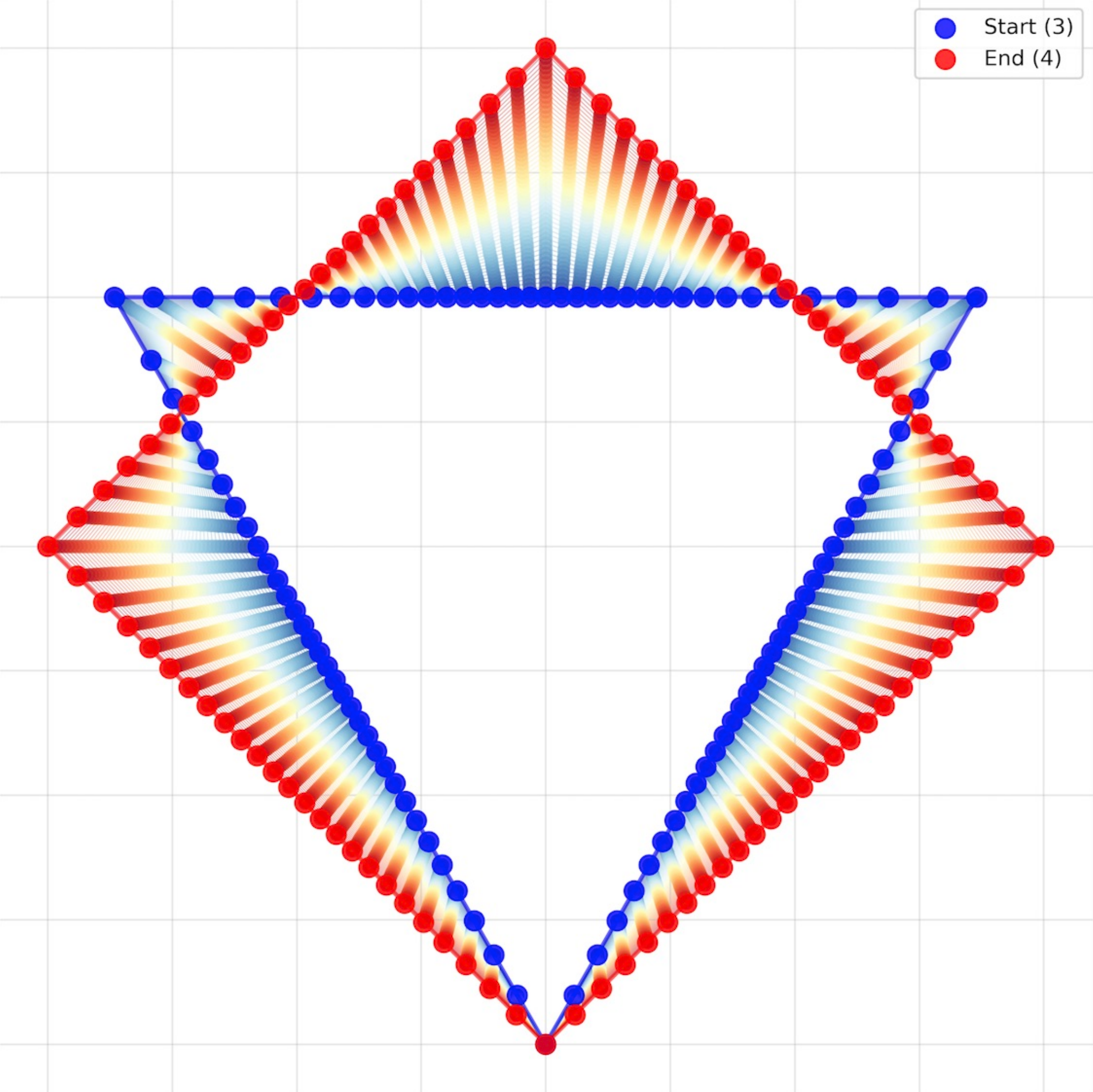}
\caption{Example of how a shape morphs from a triangle archetype to a square archetype}
\label{fig:morphing}
\end{figure}

\begin{figure}[!t]
\centering

\begin{center}\includegraphics[width=1\linewidth]{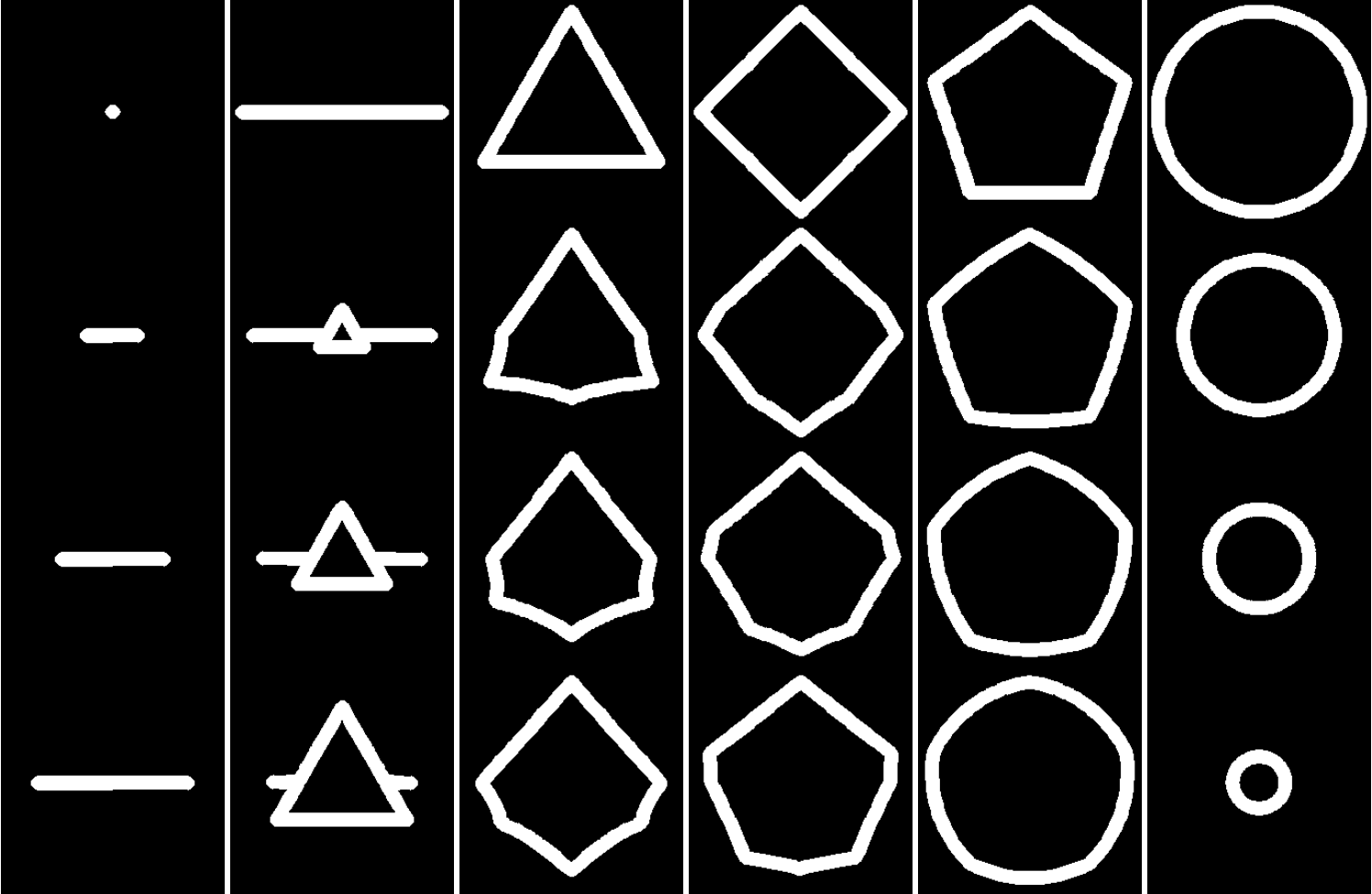} \end{center}
\caption{Examples of the data sampled from the polygonal sequence}
\label{fig:example-poly-seq}
\end{figure}

Figure \ref{fig:example-poly-seq} shows a sample of this dataset. Along
the top row are six archetype shapes from the polygonal sequence. These
shapes have a natural rank ordering: 1=point, 2=line, 3=triangle,
4=square, 5=pentagon, 6=circle (infinite sides). Transformations are
made to each shape to ``morph'' it into the next shape. This is
accomplished by moving each point radially from its position in one
shape to its position in the next. A 100\% morph refers to the complete
transformation from one shape to the next. Figure \ref{fig:morphing}
shows the path that each point takes as a triangle morphs into a square.
Along the columns of Figure \ref{fig:example-poly-seq}, the 25\%, 50\%,
and 75\% steps along each transformation are shown. To make this
sequence a complete loop, the circle transitions into the point,
resetting the process. Samples are taken evenly across this sequence for
a total of 312 images.

Using these transformed images, a classification task is set up where
each image is labeled as the nearest archetype shape, with the 50\%
transformation rounding down to the lower order. In Figure
\ref{fig:example-poly-seq}, since the first and second rows show a 25\%
and 50\% transformation respectively, the examples are associated with
the previous archetype, while the fourth row represents a 75\%
transformation and is associated with the next shape.

To classify the images, five simple convolutional neural nets (CNNs) are
trained on different subsets of the data. Each CNN has three 2D
convolutional layers with a kernel size of 3, padding of 1, and 32, 64,
and 128 dimensions respectively. These are followed by 2x2 max pooling
layers, a 128 dimensional fully connected layer with a ReLU activation
function, and a 64 dimensional fully connected layer with softmax
activation. The embeddings are pulled from the first fully connected
layer. The models are trained for 15 epochs with a 0.0001 weight decay,
0.001 learning rate, 0.25 dropout, cross-entropy loss, an Adam
optimizer, a batch size of 32, and a 70/10/20 train/validation/test
split of the input data.

Each model has a different input dataset with an increasingly large and
more representative sample of the whole dataset. These subsets are
defined by ``buffers'' around the archetype shapes. A 10\% buffer means
all the images within a 10\% transformation in either direction (i.e.,
10\% towards the lower order shape and 10\% towards the higher order
shape). Five models are trained with 10\%, 20\%, 30\%, 40\%, and 50\%
buffers. While the training data is restricted to these subsets, the
embeddings are generated using the entire dataset. This means a set of
models is created with varying levels of information about the real
domain. Models with a low buffer have a small dataset of clear shape
examples, while the 50\% buffer model sees the entire dataset during
training. This dataset restriction is meant to replicate real world
situations where a fully representative sample may not be available or
is prohibitively costly to collect. Table \ref{tab:poly-test} shows that
the models are trained to near saturation on their test sets, and that a
method is needed for model selection that does not rely on outcome
reasoning.

\begin{singlespace}
\begin{table}[H]
\centering
\begin{table}[H]
\centering
\begin{tabular}{cc}
\toprule
Buffer (\%) & Test Accuracy (\%)\\
\midrule
\cellcolor{gray!10}{10} & \cellcolor{gray!10}{100}\\
\addlinespace\addlinespace
20 & 100\\
\addlinespace\addlinespace
\cellcolor{gray!10}{30} & \cellcolor{gray!10}{100}\\
\addlinespace\addlinespace
40 & 100\\
\addlinespace\addlinespace
\cellcolor{gray!10}{50} & \cellcolor{gray!10}{96.83}\\
\bottomrule
\end{tabular}
\end{table}
\caption{Test accuracy across polygonal models}
\label{tab:poly-test}
\end{table}
\end{singlespace}

\begin{figure*}[!t]
\centering

\begin{center}\includegraphics[width=1\linewidth]{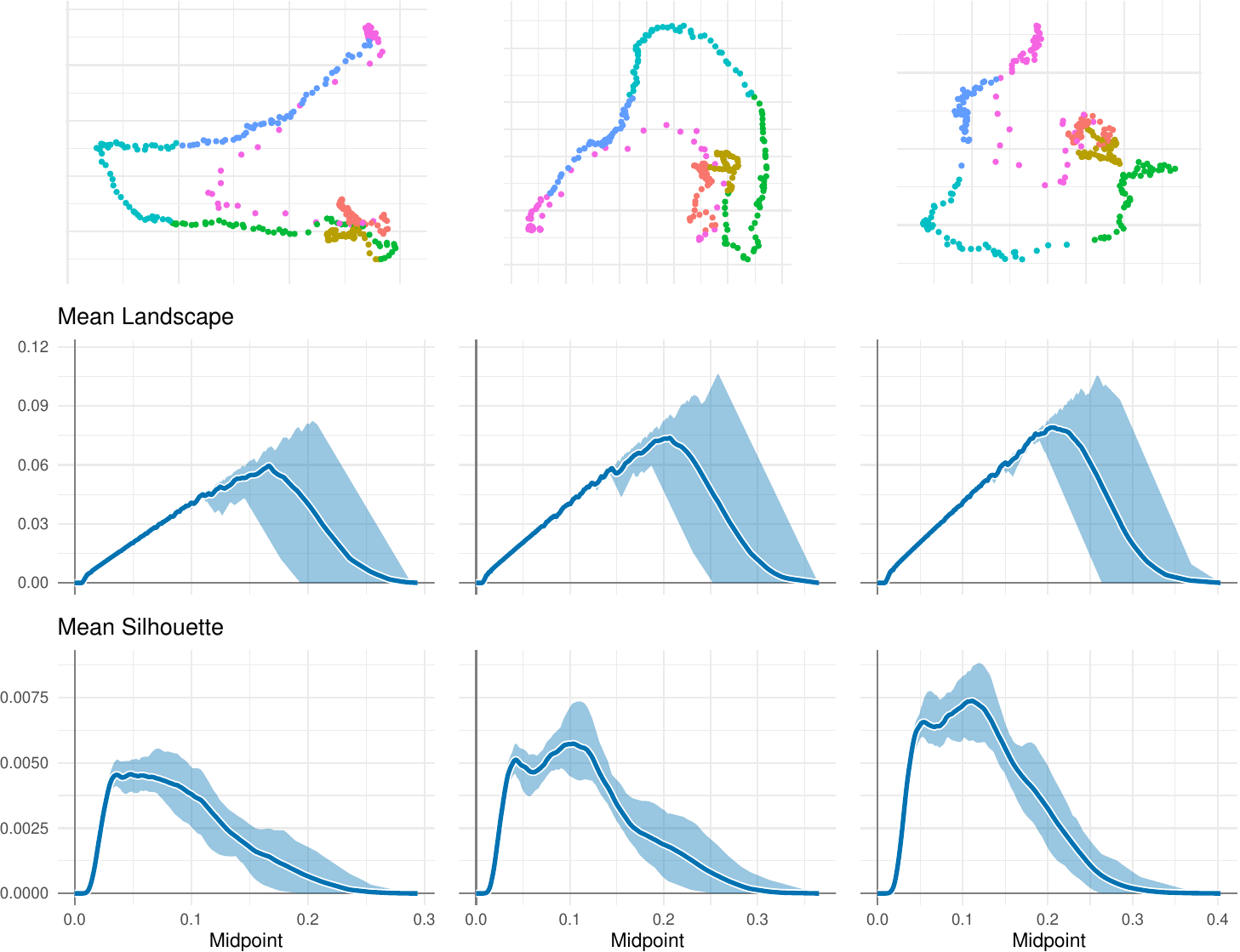} \end{center}
\caption{Examples of the embeddings and mean representations from the 10\%, 30\%, and 50\% buffer models (left to right)}
\label{fig:buffer-embeds}
\end{figure*}

\begin{figure*}[p]
\centering

\begin{center}\includegraphics[width=0.92\linewidth]{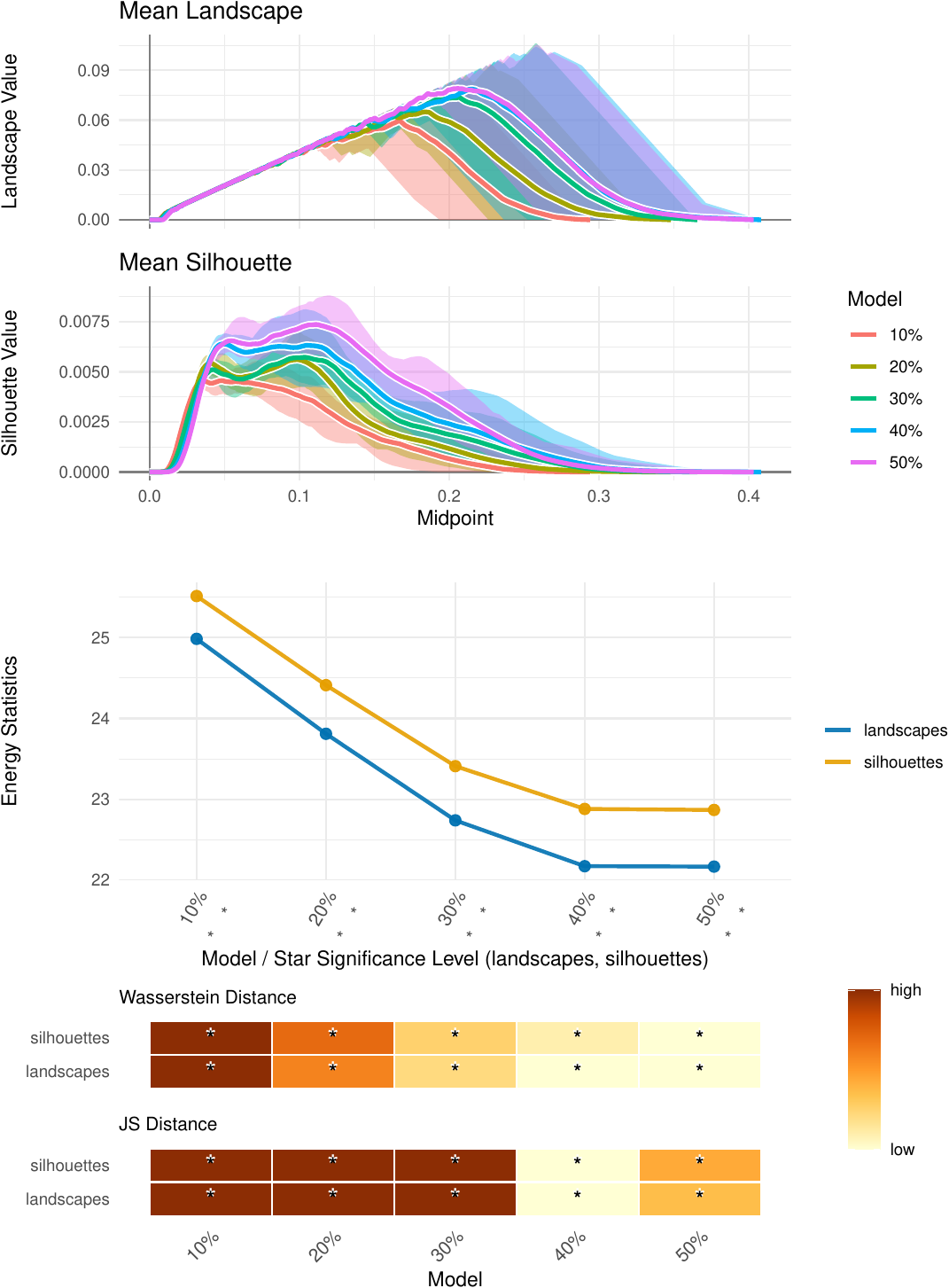} \end{center}
\caption{Comparisons of the buffer model embeddings to an equidistant embedding}
\label{fig:poly-stack}
\end{figure*}

Figure \ref{fig:poly-stack} shows the result of comparing each model to
an equidistant baseline as described in the Art results. The energy
statistics show a clear progression in the embedding space as the models
are trained on more data. Figure \ref{fig:buffer-embeds} demonstrates
what this transition looks like in the embeddings. The high dimensional
embeddings are plotted using PCA for the 10\%, 30\%, and 50\% buffer
models and the associated mean representations are shown below the plot.
The 10\% model embeddings form smooth transitions between the shape
archetypes while the 50\% embeddings have moved into tighter clusters.
The 10\% model correctly learns that the dataset comes from a smooth
transition of shapes. Since that model was trained on only clear
examples it appears to have learned that similar shape transformations
should be mapped to similar locations in the embedding space. This is
fantastic if understanding shape similarities is the model's goal, but
this is a classification task, clear clusters of classes is the goal.
Rather than a loop representing smooth transitions, a separated semantic
space indicates a stronger understanding of class labels. In the 50\%
model, the embedding space shows tighter clustering of the shapes. While
the edge cases are still floating between clusters, the concept
coherence for each archetype is much higher.

\begin{figure*}[!t]
\centering

\begin{center}\includegraphics[width=1\linewidth]{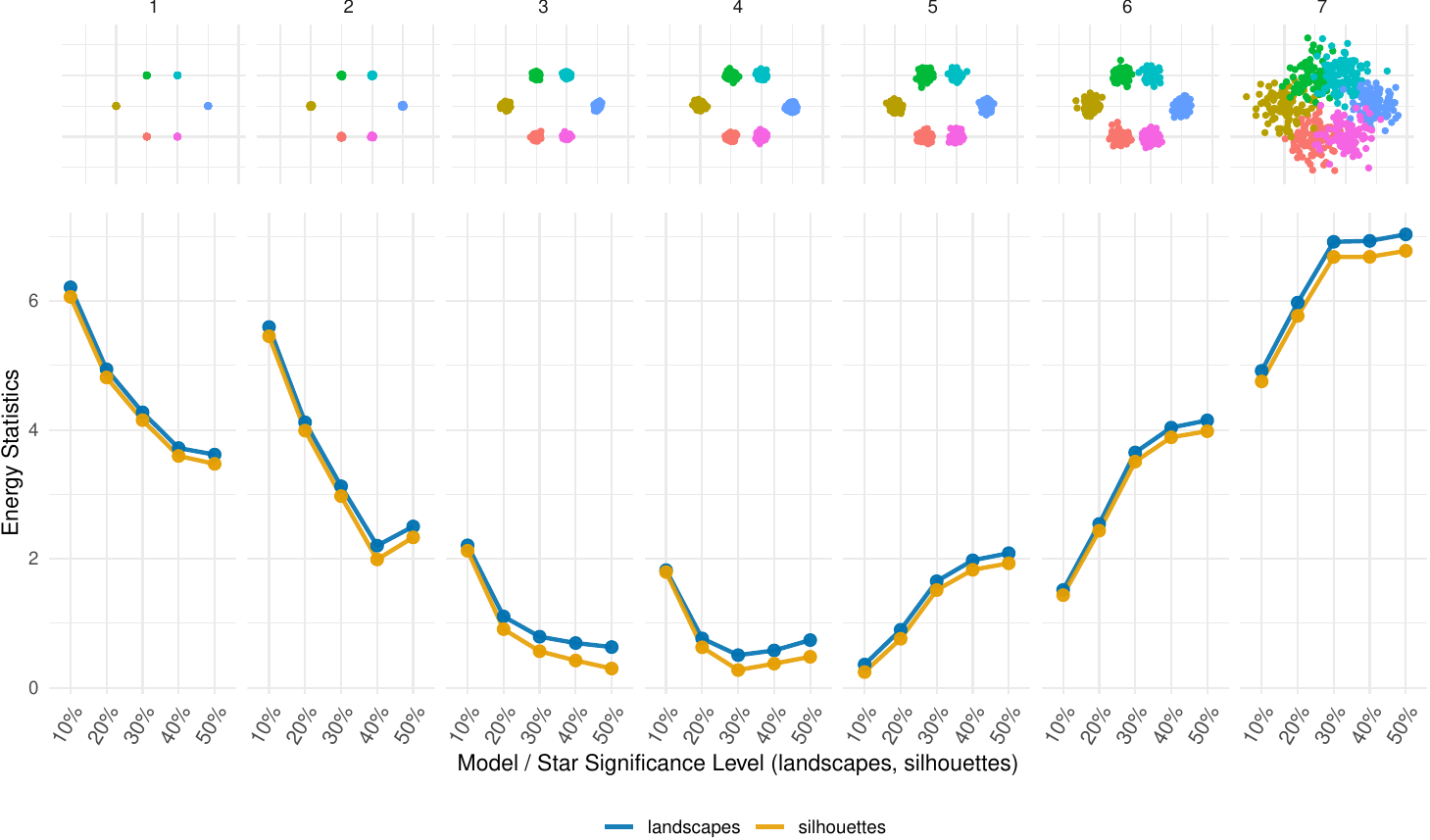} \end{center}
\caption{Example of model selection by checking the alignment of the buffer models to curated preferences}
\label{fig:poly-model-sel}
\end{figure*}

The preference for semantic separation is only a preference though. It
may be the case that tightly clustered concepts is the type of alignment
developers are trying to teach the model. It may instead be the case
that given a few models which score equally well on the test set, an
internal representation that aligns with the true data generation
process is preferred. Figure \ref{fig:poly-model-sel} shows how this
model selection can be handled across different preferences.

The top row shows seven plots with clusters representing the six
archetype shapes. On the left, the points are tightly clustered and to
the right the points are increasingly noisy. As the noise increases, the
groups begin to blur together, eventually forming a ring shape. Through
this plot, preferences for these semantic structures have been encoded
into intuitive 2D representations. Below these plots are the energy
statistics associated with each model compared to the top point cloud.
The graphs are showing each model's alignment with a curated and easily
definable knowledge structure.

If a developer prefers a model that has learned distinct clusters of
concepts, then the left most plot shows the 50\% model to be most
aligned with this preference. If a developer prefers a smooth transition
between concepts, the right plot shows that the 10\% model is most
aligned. If the developer prefers a semantic structure in the middle,
when the concepts are distinct but blurry, most of the models are
equally aligned.

By rigorously testing the alignment between embedding spaces of
arbitrary dimensionality, the tools allow for model selection based on
preferences that are defined intuitively or come from more formalized
encodings like ontologies.

\begin{center}\rule{0.5\linewidth}{0.5pt}\end{center}

\section{Conclusion}\label{conclusion}

This work fills a gap in the testing frameworks of modern AI
development. These methods are not meant to replace outcome reasoning,
rather they are meant to supplement existing methods. Combining these
techniques with outcome reasoning can give a more holistic view of a
model's performance.

The main benefit of this method is the development of flexible
baselines. Using baselines from 2D plots, where developers can visually
plot intuitive relationships between concepts, lowers the barrier of
entry for testing alignment and preferences of non-technical audiences
can be encoded quickly and flexibly. Alternatively, graphs can be
created from formal ontologies or knowledge encodings to be used as
baselines, which is especially useful in domains with a strict hierarchy
of rules or classifications like law or policy.

Since these representations are agnostic to the embedding dimensionality
or modality, different model architectures can be compared consistently.
By applying adaptations and training models from scratch, these results
show how this technique tracks changes in semantic structure and how
conceptual coherence can be induced, optimized, or degraded.

Persistence landscapes and silhouettes offer a way to characterize
global and local structure that focuses on connectivity and shape rather
than scale. These results show that topological representations can
effectively aid in model selection and AI alignment even when test sets
become saturated.

\section{Future Work}\label{future-work}

This work demonstrates the utility of the topology based semantic tests
and opens the possibility of future work that connects these methods to
diffusion models and an end-to-end decision theoretic framework via
possibility theory.

\subsection{Synthetic Data}\label{synthetic-data}

Synthetic data generation may be seen as a solution to the data sampling
problem, but it may be ineffective.

If the goal is to generate similar data samples to the existing dataset,
the persistence landscapes would likely remain similar since the
clusters would remain largely the same all new points would appear as
low persistence features. The silhouettes may capture the change in the
count of low persistence features since they average the features.
Comparing the landscapes and silhouettes in this situation may offer
insights to data sampling strategies.

However, if the goal of synthetic data generation is to sample new
spaces or increase coverage of the domain, this could hurt the
representation's ability to differentiate models. If data is
systematically generated to fill the gaps between existing data points,
then the structure of the embedded data may start to degrade. Since the
topological approaches focus on connectivity, if data results in a
uniform semantic structure, the topological representations will become
less distinguishable. The issue of representativeness is well known in
synthetic data generation but may be exacerbated if the models have
entered an interpolative regime as suggested by the double descent work.

This is an interesting line of future work because if the
representations do change when synthetic data is introduced, then the
technique can be used to identify datasets generated synthetically.

\subsection{Data Identifiability}\label{data-identifiability}

These topological approaches capture how point clouds connect as
filtrations are applied, but they ignore which data points are
connecting. If two points are swapped in the embedded space, the
representations will not change at all since the data points are not
tracked in the persistence diagrams directly.

However, adding a dimension to the persistence diagram and mean
representations may keep them connected to the original data points. For
example, a cylindrical coordinate system can be used where radius and
height correspond to the original birth and death coordinates. Then by
partitioning the polar plane into sections associated with each data
point a three dimensional representation is created that maintains a
connection to the original data.

\subsection{Connection to Diffusion
Models}\label{connection-to-diffusion-models}

The topological methods use filtrations to build persistence diagrams,
but filtrations can also be used to build diffusions models. Rather than
building the topological landscapes and silhouettes, fitting diffusion
models to the embedded data may offer a different but useful view of the
embeddings. Diffusion models are well studied and well understood. The
existing methods of comparing diffusion models may be a promising
non-topological route to alignment checks.

\subsection{Rotation Transformation and Possibility
Theory}\label{rotation-transformation-and-possibility-theory}

Adding an extra rotation to the transformed persistence diagram is not
the only nonstandard transformations that can be applied. The small
extra rotation is used in this work since it remains closely aligned
with standard application but results in more descriptive landscapes.

More flexibly, the generation of landscapes and silhouettes can be seen
as an aggregation of many convolutions of input functions around each
persistence point and a filter function. This is an explicit connection
in the literature between persistence silhouettes, persistence images,
and kernel density estimation. However, landscapes with any rotation or
rescaling can also be created using the right boxcar input and filter
functions.

The main differences between these two representations is the
aggregation method. Persistence landscapes use a maxitive kernel while
the silhouettes use a summative kernel. Summative kernels are common in
probability theory, but maxitive kernels are more common in possibility
theory. Possibility theory is a branch of statistics related to
probability theory, but instead of relying on classical set theory and
binary logic, possibility theory relies on fuzzy set theory and
multi-valued logic. Possibility theory also has strong connections to
multi-criteria decision making (MCDM). By connecting these topological
methods to formal decision making frameworks, a pipeline for model
selection and alignment can be defined from datasets to deployment which
has expert knowledge encoded into the process and a consistent
underlying mathematical backbone.

\section*{Acknowledgements}\label{acknowledgements}
\addcontentsline{toc}{section}{Acknowledgements}

The authors are especially grateful to Karen Kafadar, whose guidance as
the first author's dissertation co-advisor shaped this work throughout.
The authors also thank the members of the dissertation committee, Anette
(Peko) Hosoi, Taylor Brown, and David Evans, for their feedback and
support.

\printbibliography

\end{document}